\pdfoutput=1
\documentclass[11pt]{article}

\usepackage[final]{acl}
\usepackage{amsfonts}
\usepackage{times}
\usepackage{latexsym}

\usepackage[T1]{fontenc}
\usepackage[utf8]{inputenc}

\usepackage{microtype}

\usepackage{inconsolata}

\usepackage{graphicx}
\usepackage{subcaption}

\usepackage{hyperref}
\usepackage{underscore}
\usepackage{booktabs}   % 表格
\usepackage{tabularx}   % 表格
\usepackage{siunitx}    % 表格标准差下标
\usepackage{makecell}   % 三线表-竖线
\usepackage{multirow}
\usepackage{colortbl}
\usepackage{tcolorbox}
\usepackage{amsmath}    % 对齐
\usepackage{algorithm}
\usepackage{algpseudocode}
\usepackage{CJKutf8}
\usepackage{tcolorbox}
\usepackage{verbatim}
\usepackage{subfiles} % Best loaded last in the preamble
\usepackage{verbatim}
\usepackage[utf8]{inputenc}
\tcbuselibrary{skins,breakable} % Add breakable library for multi-page boxes if needed
\usepackage{pifont}
\usepackage{amssymb}
\newcommand{\smaller}{\textcolor[HTML]{3b6291}}
\newcommand{\bigger}{\textcolor[HTML]{943c39}}

\newif\ifshowrev
\showrevfalse

\ifshowrev
  \newcommand{\rev}[1]{\textcolor{blue}{#1}}
\else
  \newcommand{\rev}[1]{#1}
\fi

\newcommand{\Model}{\textbf{DyReMe}}
\newcommand{\Gen}{\texttt{DyGen}}
\newcommand{\Eval}{\texttt{EvalMed}}
\newcommand{\adjval}[1]{\makebox[3em][r]{(1)}}

\title{Inflated Excellence or True Performance? Rethinking Medical Diagnostic Benchmarks with Dynamic Evaluation}
\author{
 \textbf{Xiangxu Zhang\textsuperscript{1}},
 \textbf{Lei Li\textsuperscript{1}},
 \textbf{Yanyun Zhou\textsuperscript{1}},
 \textbf{Xiao Zhou\textsuperscript{1,3,4}}\thanks{Corresponding Author},
\\
 \textbf{Yingying Zhang\textsuperscript{2}},
 \textbf{Xian Wu\textsuperscript{2}}\footnotemark[1]
\\
 \textsuperscript{1}GSAI, Renmin University of China,
 \textsuperscript{2}Tencent Jarvis Lab,
\\
 \textsuperscript{3}\rev{Beijing Key Laboratory of Research on Large Models and Intelligent Governance,}
\\
 \textsuperscript{4}\rev{Engineering Research Center of Next-Generation Intelligent Search and Recommendation, MOE}
\\
 \texttt{\{xansar,xiaozhou\}@ruc.edu.cn}
\quad
 \texttt{kevinxwu@tencent.com}
}

\begin{document}
% \subfile{sections/revision_explanations}

\maketitle

\begin{abstract}
Medical diagnostics is a high-stakes and complex domain that is critical to patient care. 
% However, current evaluations of large language models (LLMs) are fundamentally misaligned with real-world clinical practice. 
\rev{However, current evaluations of large language models (LLMs) remain limited in capturing key challenges of clinical diagnostic scenarios.}
{Most rely on benchmarks derived from public exams, raising contamination bias that can inflate performance, and they overlook the confounded nature of real consultations beyond textbook cases.}
% {Recent dynamic evaluations offer a promising alternative, but often remain insufficient for diagnosis-oriented benchmarking, with limited coverage of clinically grounded confounders and trustworthiness beyond accuracy.}
{Recent dynamic evaluations offer a promising alternative, but often remain insufficient for \rev{diagnosis-oriented} benchmarking, with limited coverage of clinically grounded confounders and trustworthiness beyond accuracy.}
% To address these gaps, we propose \Model, a dynamic benchmark for medical diagnostics that better reflects real clinical practice. 
To address these gaps, we propose \Model, a dynamic benchmark for medical diagnostics that \rev{provides a controlled and scalable stress test of diagnostic robustness.}
% Unlike static exam-style questions, \Model\ generates fresh, consultation-like cases that introduce distractors such as differential diagnoses and common misdiagnosis factors. 
Unlike static exam-style questions, \Model\ generates fresh, consultation-\rev{style} cases that incorporate \rev{clinically grounded confounders}, such as differential diagnoses and common misdiagnosis factors.
% It also varies expression styles to mimic diverse real-world query habits. 
It also varies expression styles to \rev{capture heterogeneous patient-style descriptions}.
Beyond accuracy, \Model\ evaluates LLMs on three additional clinically relevant dimensions: veracity, helpfulness, and consistency. 
% Our experiments demonstrate that this dynamic approach yields more challenging and realistic assessments, revealing significant misalignments between the performance of state-of-the-art LLMs and real clinical practice. 
Our experiments show that this dynamic approach yields more challenging assessments and \rev{exposes substantial weaknesses of state-of-the-art LLMs under clinically confounded diagnostic settings.}
% These findings highlight the urgent need for evaluation frameworks that better reflect the demands of trustworthy medical diagnostics.
These findings highlight the urgent need for evaluation frameworks that better \rev{assess trustworthy medical diagnostics under clinically grounded confounders.}
\footnote{We have officially released the code \href{https://github.com/xansar/dyreme}{here}. \rev{To support continual evaluation, we plan to refresh \Model\ every six months by updating benchmark instances and releasing versioned snapshots.}}

\end{abstract}

    \section{Introduction}
    \label{sec:intro}
    Accurate medical diagnostics is vital for patient health and effective treatment. However, the inherent complexity (e.g., the diversity of symptom-diagnosis relationships) and external factors (e.g., incomplete information and various expression styles) pose significant challenges~\cite{studdertDefensiveMedicineHighRisk2005,naikLegalEthicalConsideration2022}.
    Misdiagnoses can lead to severe consequences, such as higher mortality rates~\cite{pourafkariFrequencyInitialMisdiagnosis2017} and increased healthcare burdens~\cite{juarez2006costs}, 
    % underscoring the need for trustworthy diagnostic tools aligned with clinical practice~\cite{paiExplainableAnalyticsExploring2024}. 
    underscoring the need for trustworthy diagnostic tools~\cite{paiExplainableAnalyticsExploring2024}. 
    
    \begin{figure}[t]
        \centering
        \includegraphics[width=\linewidth]{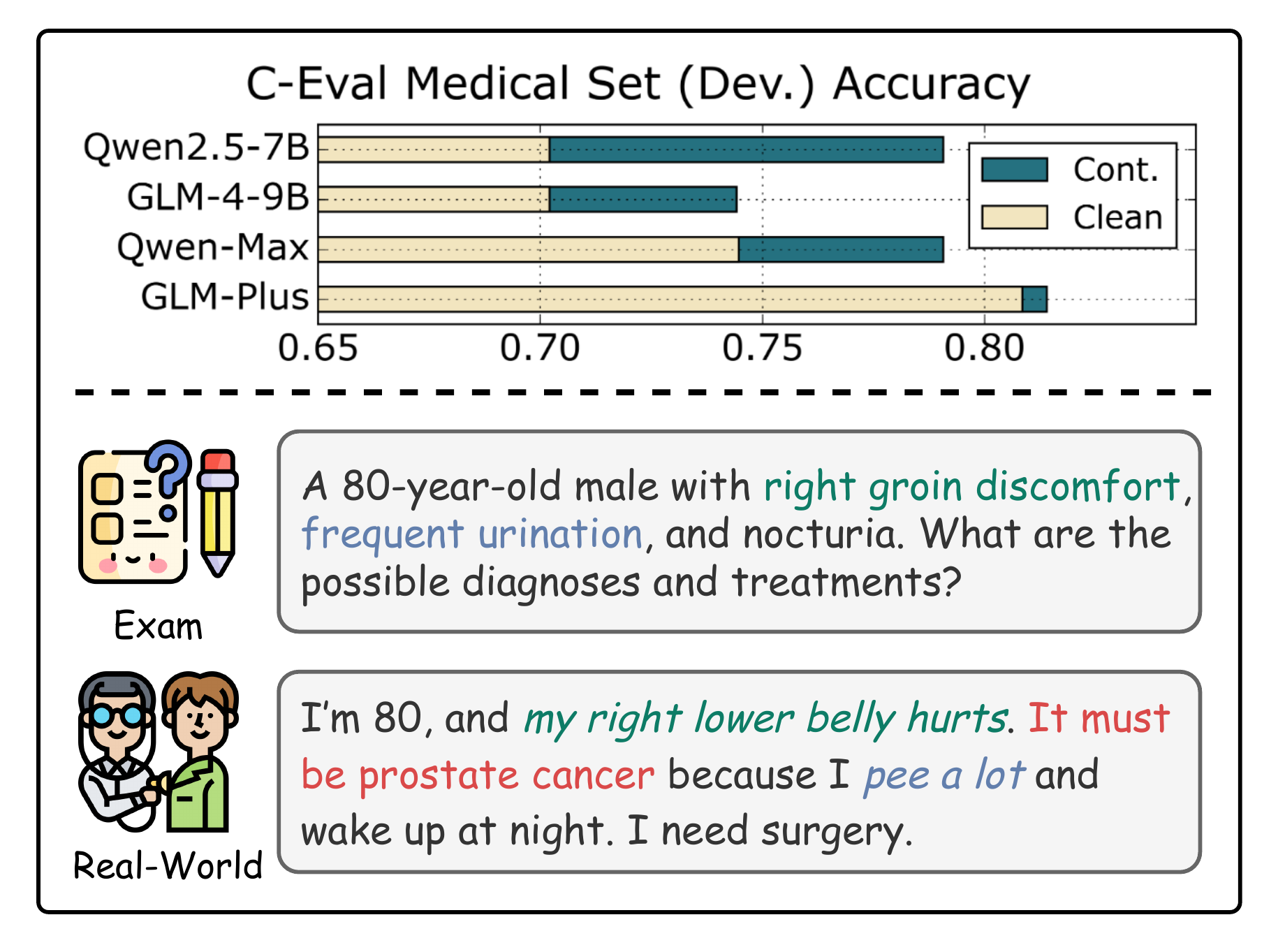}
        \caption{% (Top) Performance on C-Eval medical development set.
        (Top) Contamination data experiments on C-Eval~\cite{huangCEvalMultiLevelMultiDiscipline2023, liOpenSourceData2024a}.
        % The better results on contaminated data suggest a possible overestimation of model performance.
        (Bottom) Unlike standardized patients in current benchmarks, real-world patients express symptoms unprofessionally and may misdiagnose themselves, misleading doctors.}
        \label{fig:data-contamination}
    \end{figure}
    
    Recent advances in artificial intelligence (AI)~\cite{li2025fap,he2025motifgpl,li2025spatio,hong2025lost} and large language models (LLMs)~\cite{yong2025motivebench,zhu2025mohobench,guo2025counterfactual,yongthink,yong2026intelli} show promise in assisting healthcare professionals by processing medical knowledge~\cite{chenHuatuoGPTIIOnestageTraining2024,zhang2026r2medbenchmarkreasoningdrivenmedical,zhu2025hypertension} and supporting diagnostic decisions~\cite{chenCoDInterpretableMedical2024a, zhang2026hypemed}. These models can analyze clinical cases~\cite{wangCMBComprehensiveMedical2024}, identify patterns in laboratory medicine data~\cite{ullahChallengesBarriersUsing2024}, potentially improve diagnostic accuracy~\cite{ fanAIHospitalBenchmarking2024a}, and perform multiple tasks~\cite{zhu2025hypertension,hong2025lost,zhou2025tricolore,li2025leave,zhang2026social}. 
    To assess their capabilities, \emph{static} benchmarks based on medical exams have been developed~\cite{liuBenchmarkingLargeLanguage2023a,caiMedBenchLargeScaleChinese2024}, where the test questions remain fixed across models and time~\cite{zhuDynamicEvaluationLarge2024, wangBenchmarkSelfEvolvingMultiAgent2024,hong2026wed,guo2026not,guo2025sorex}.

    % However, a key question remains: \emph{Do static benchmarks reflect real-world diagnostic capabilities of LLMs?}~\cite{jiangInvestigatingDataContamination2024,guptaImprovingModelEvaluation2024}.
    However, a key question remains: \emph{Do static benchmarks reflect \rev{true medical} diagnostic capabilities of LLMs?}~\cite{jiangInvestigatingDataContamination2024,guptaImprovingModelEvaluation2024}.
    {Two limitations are particularly relevant:
    1) \underline{\rev{Biased estimation of diagnositc capability}}: Because many medical benchmarks are \emph{public and static}, they are vulnerable to \emph{data contamination}~\cite{liOpenSourceDataContamination2024,xuBenchmarkDataContamination2024}.
    This overlap biases evaluation: high scores may reflect exposure rather than generalizable reasoning, overestimating real-world performance.
    2) \underline{Misalignment with real-world scenarios}~\cite{liu2024medchain,park2024assessing}: 
    Even when contamination is controlled, exam-style benchmarks adopt standardized, well-formed case descriptions and an accuracy-centric protocol, 
    % whereas real-world queries are often incomplete, lay, and confounded by self-diagnosis, which can mislead clinical decision-making~\cite{graberDiagnosticErrorInternal2005,normanCausesErrorsClinical2017}.
    whereas \rev{paitent} queries are often incomplete, lay, and confounded by self-diagnosis, which can mislead clinical decision-making~\cite{graberDiagnosticErrorInternal2005,normanCausesErrorsClinical2017}.
    }
    
    Fig.\ref{fig:data-contamination} further illustrates this gap: the top panel shows contamination-induced inflation, while the bottom panel highlights patient-style expressions and diagnostic confounders absent from exam questions.
    % These issues call for an evaluation approach that mitigates contamination and better aligns with real-world diagnostic interactions, without relying on costly annotation~\cite{lyuRJUAQAComprehensiveQA2024a}.
    These issues call for an evaluation approach that mitigates contamination and better \rev{captures clinically plausible confounders in diagnosis-oriented queries}, without relying on costly annotation~\cite{lyuRJUAQAComprehensiveQA2024a}.

    One natural direction is \emph{dynamic} evaluation, which generates new test cases by transforming existing questions (e.g., rewriting~\cite{wangBenchmarkSelfEvolvingMultiAgent2024}, adding noise~\cite{zhuDynamicEvaluationLarge2024}, or altering graphs~\cite{zhuDyValDynamicEvaluation2023}) to reduce direct overlap with public benchmarks.
    However, such transformations are often surface-level and largely preserve the underlying clinical setting, leaving real-world misalignment unresolved in diagnostic confounders~\cite{graberDiagnosticErrorInternal2005} and patient-style expressions. Moreover, it still does not change the sole focus on accuracy.
    Consequently, dynamic evaluation remains insufficient for benchmarking LLMs under realistic diagnostic conditions. This motivates an evaluation framework that 
    % (i) generates \emph{unseen} cases \emph{with realistic confounders} 
    (i) generates \emph{unseen} cases \emph{with \rev{clinically grounded} confounders} 
    and (ii) evaluates \emph{trustworthiness beyond accuracy}.

    To address both contamination-induced evaluation bias and real-world misalignment, we propose \Model, a \underline{\textbf{Dy}}namic, \rev{\underline{\textbf{Re}}aliable} evaluation framework for \underline{\textbf{Me}}dical diagnostics. \Model~has two key components: \Gen~and \Eval.
    \Gen~generates realistic diagnostic benchmarks by creating \emph{unseen} questions that incorporate differential diagnoses and misdiagnosis factors as diagnostic distractors~(e.g., a \emph{sinusitis} patient \underline{self-diagnosed} with \emph{periodontitis}). 
    Additionally, \Gen~takes patient expression features into account, 
    ensuring that the generated questions reflect real-world query styles 
    (e.g., a patient describing \emph{radiating pain} as ``\emph{an electric shock running down my arm}'').
    % Together, these designs explicitly inject real-world confounders and patient-style expressions, reducing the misalignment between benchmarks and practical diagnostic interactions.
    Together, these designs introduce clinically plausible confounders and patient-style expressions, \rev{enabling controlled stress tests for clinically plausible failure modes.}
    \Eval~assesses four aspects: accuracy, veracity, helpfulness, and consistency. It checks whether the model gives the correct diagnosis, corrects health rumors, provides useful explanations, and gives stable answers. By moving beyond an accuracy-only protocol, \Eval~addresses evaluation-protocol misalignment and reduces over-optimistic conclusions drawn from static, exam-style benchmarks.
    Experiments demonstrate that \Model~not only generates unseen, confounder-rich diagnostic cases (\Gen) but also provides trustworthy, human-aligned evaluations (\Eval), 
    % revealing limitations of LLMs under realistic diagnostic conditions.
    revealing limitations of LLMs under \rev{challenging} diagnostic conditions.

    \noindent
    \textbf{Contributions.} 
    (1)~We propose \Model, a dynamic evaluation framework for medical diagnostics that combines benchmark generation (\Gen) with trustworthy evaluation (\Eval). 
    (2)~Using \Gen, we construct a challenging benchmark that reflects key clinical confounders.
    (3)~We evaluate 12 LLMs with \Eval, uncovering limitations in their diagnostic capabilities.

    \section{Related Work}
    \begin{figure*}[htbp]
        \centering
        \includegraphics[width=\linewidth]{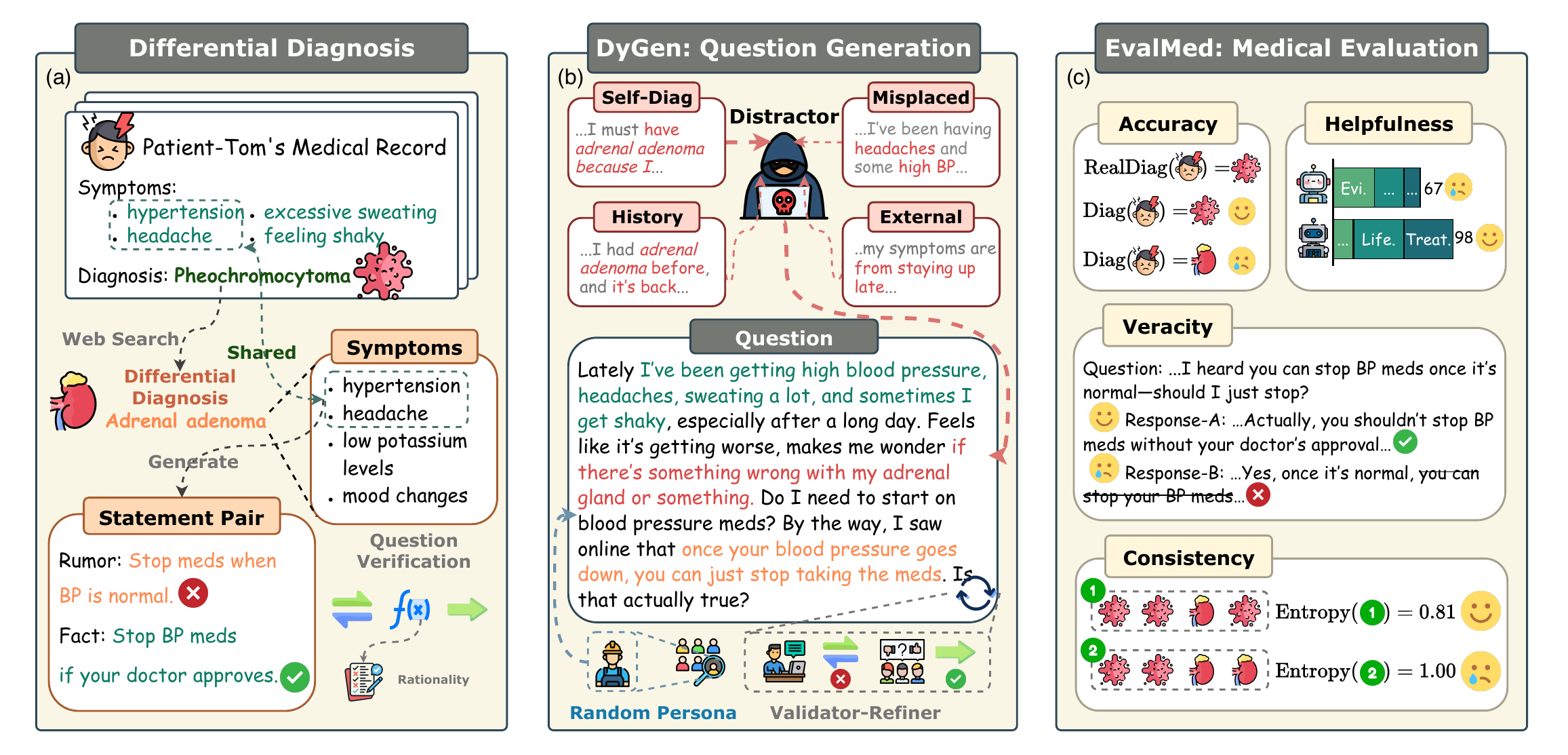}
        \caption{Overview of \Model. (a) Differential diagnosis construction and medical rumor generation. (b) Question generation with trap selection, persona style, and refinement. We focus on \textbf{Chinese} questions and the example question is translated from Chinese. (c) \Eval\ assesses Accuracy, Veracity, Helpfulness, and Consistency.}
        \label{fig:framework}
    \end{figure*}

    \paragraph{Static Benchmarks for Medical LLMs.}
    Medical LLMs are often evaluated on benchmarks such as PubmedQA~\cite{jinPubMedQADatasetBiomedical2019}, MedQA~\cite{jin2021disease}, CMB~\cite{wangCMBComprehensiveMedical2024}, and CMExam~\cite{liuBenchmarkingLargeLanguage2023a}, which come from static medical exams.
    Although models like GPT-4o~\cite{openaiGPT4oSystemCard2024} perform well on such benchmarks, their public availability raises concerns about contamination and inflated scores.
    Moreover, exam-style questions often fail to capture real clinical complexity.
    Newer datasets such as MedBench~\cite{caiMedBenchLargeScaleChinese2024}, DxBench~\cite{chenCoDInterpretableMedical2024a}, CliMedBench~\cite{ouyangCliMedBenchLargeScaleChinese2024}, and others~\cite{
    nessMedFuzzExploringRobustness2024,
    kimMedExQAMedicalQuestion2024,
    zhangLLMEvalMedRealworldClinical2025, bediMedHELMHolisticEvaluation2025,pan2025beyond} incorporate expert annotations and better reflect clinical reasoning (e.g., RJUA~\cite{lyuRJUAQAComprehensiveQA2024a}), but remain costly, prone to contamination, and limited in representativeness. {Thus, despite higher reasoning fidelity, they do not scale or refresh fast enough to match the pace of medical LLMs.}

    \paragraph{Dynamic Evaluation.}
    Dynamic evaluation has been proposed to address the limitations of static benchmarks by rewriting, perturbing, or paraphrasing raw questions~\cite{kielaDynabenchRethinkingBenchmarking2021, 
    zhangDARGDynamicEvaluation2024, 
    whiteLiveBenchChallengingContaminationFree2024, 
    wangSOTOPIA$pi$InteractiveLearning2024}.
    In this work, we focus on generating new medical diagnostic benchmarks from seed cases.
    Among these approaches, dynamic evaluation methods based on LLMs, such as DyVal2~\cite{zhuDynamicEvaluationLarge2024} and Self-Evolving~\cite{wangBenchmarkSelfEvolvingMultiAgent2024} employ LLMs as generators to improve scalability and adaptability. DyVal2 introduces ``probing'' and ``judging'' agents to create new questions through paraphrasing, adding noise, and permutation. Self-Evolving applies six different reframing operations to construct evolving instances, testing LLMs across query variations, noise, and problem-solving robustness.
    However, these methods overlook the specific requirements of medical diagnostics and fail to address the limitations of static benchmarks, as they mainly rely on superficial transformations. 
    \rev{Recent medical evaluation frameworks such as CRAFT-MD~\cite{johri2024craft}, MedKGEVal~\cite{yu2025medkgeval}, and ClinDEF~\cite{tang2025clindef} further explore multi-turn, knowledge-grounded, or workflow-oriented diagnostic evaluation. These frameworks are complementary to \Model. While they emphasize richer clinical interactions and workflow-oriented assessment, \Model\ focuses on controlled and scalable diagnosis-oriented stress testing under clinically grounded confounders.}

    \paragraph{Summary.} 
    Existing benchmarks overestimate performance and fail to reflect clinical practice. Dynamic methods attempt to bridge this gap but still miss real-world distractors and trustworthy assessment. To address this gap, we introduce \Model.

    \section{Methodology}
    \label{sec:methodology}
   We propose \Model~(Fig.\ref{fig:framework}), a dynamic evaluation framework that better aligns with real-world clinical practice. It comprises two components: (1)~\Gen, a generation module that creates realistic and challenging questions, and {(2)~\Eval, an evaluation module that assesses LLMs across four clinically relevant dimensions: \emph{accuracy}, \emph{veracity}, \emph{helpfulness}, and \emph{consistency}.}
    We begin by formally defining the diagnostic task. Given a question $q$ based on a set of symptoms $S = \{s_i\}_{i=1}^N$ and an underlying diagnosis $d$, the LLM ${M}$, parameterized by $\theta$, generates a response $\hat{a} = {M}(q)$. A scoring function $f$ is used to assess the quality of $\hat{a}$.

    \subsection{\Gen: Dynamic Generation}
    \label{sec:trap-questions-generator}
    Current diagnostic benchmarks oversimplify the complexity and variability of real-world diagnostics, resulting in a misalignment with actual clinical consultations. Trustworthy diagnosis requires consideration of differential diagnoses~\cite{arter1979differential} and misdiagnosis factors~\cite{graberDiagnosticErrorInternal2005, normanCausesErrorsClinical2017}, collectively referred to as ``diagnostic distractors.''
    To simulate real-world complexity, \Gen~first constructs diagnostic distractors by integrating differential diagnoses and misdiagnosis factors. It then incorporates these distractors into the questions and rephrases them to match authentic clinical expressions. Finally, \Gen~refines the questions using a validator-refiner iterative loop to ensure clinical validity and maintain real-world complexity.

    \paragraph{Differential Diagnosis.}
    \label{sec:differential-diagnosis} 
    %\Gen~retrieves differential diagnoses $d_{\text{dis}}$ of the original diagnosis $d_{\text{org}}$, which shares similar symptoms, to mirror the real-world differential diagnostic workflow of physicians. This is achieved by Retrieval-Augmented
    % Generation~($\mathcal{G}_{\text{RAG}}$)~\cite{lewisRetrievalAugmentedGenerationKnowledgeIntensive2020}.
    \Gen~retrieves similar diagnoses $d_{\text{dis}}$ for a given original diagnosis $d_{\text{org}}$, mimicking real-world differential diagnosis. 
    This is achieved using Retrieval-Augmented Generation~($\mathcal{G}_{\text{RAG}}$)~\cite{lewisRetrievalAugmentedGenerationKnowledgeIntensive2020}.
    %For instance, \Gen~generates $d_{\text{dis}}$=``\textit{Adrenal adenoma}'' as the differential diagnosis of $d_{\text{org}}$=``\textit{Phoechromocytoma}'' based on medical encyclopedias,
   % because $d_{\text{dis}}$ shares common symptoms (\textit{hypertension} and \textit{headache})
    %with $d_{\text{org}}$~(Fig.\ref{fig:framework}a). Eq.\ref{eq:differential} describes the above process:
    For example, $\mathcal{G}_{\text{RAG}}$ retrieves $d_{\text{dis}}$ = ``\textit{Adrenal adenoma}'' as a differential diagnosis for $d_{\text{org}}$ = ``\textit{Pheochromocytoma}'' based on medical encyclopedias, since both share common symptoms like \textit{hypertension} and \textit{headache} (Fig.\ref{fig:framework}a). Formally, $d_{\text{dis}}=\mathcal{G}_{\text{RAG}}(d_{\text{org}}).$
    % \begin{equation}
    %     \begin{aligned}
    %                     & d_{\text{dis}}= \mathcal{G}_{\text{RAG}}(d_{\text{org}}).                      \\
    %         % \text{s.t.} & \quad \text{Sym}(d_{\text{dis}}) \cap \text{Sym}(d_{\text{org}}) \neq \emptyset,
    %     \end{aligned}
    %     \label{eq:differential}
    % \end{equation}
    % here, $\text{Sym}(\cdot)$ denotes the symptom set. 
    % $d_{\text{dis}}$ will be used with misdiagnosis factors to consturct diagnostics distractors, which are designed to align with real-world diagnostic complexity and increase the challenge of the generated questions.
    % Each diagnostic question includes these shared symptoms to increase difficulty along with distinguishing symptoms that enable logical elimination of $d_{\text{dis}}$. 
    % Additionally, patient queries are crafted to reflect concerns about $d_{\text{dis}}$, compelling LLM to engage in nuanced diagnostic reasoning. In fact, even unrelated diagnoses can also cause interference but may be less effective. This approach ensures that the generated questions are clinically challenging and relevant to real-world scenarios.

    \paragraph{Misdiagnosis Factors.}
    Anchor bias~\cite{graberDiagnosticErrorInternal2005}, posterior probability error~\cite{nendazDiagnosticErrorsFlaws2012}, distraction~\cite{graberDiagnosticErrorInternal2005}, and symptom overestimation~\cite{braun2017diagnostic} pose challenges in clinical practice.
    To reflect these factors, {\Gen~designs four diagnostic traps~(Fig.\ref{fig:framework}b): $\mathcal{S}$=\{{self-diagnosis}, {distracting history}, {external noise}, {misplaced symptoms}\}.} 
    Given a question $q_{\text{org}}$ (e.g., ``\emph{What's wrong with me if I have hypertension and headache?}''), \Gen~selects a trap $s$ and a differential diagnosis $d_{\text{dis}}$ (e.g., ``\emph{Adrenal adenoma}''), and combines them to form a misleading question $q_{\text{trap}}$ (e.g., ``\emph{\underline{I might have an adrenal adenoma} due to hypertension and headache. Can you give me medicine?}''):
    \begin{equation}
        q_{\text{trap}}= \mathcal{T}_{\text{trap}}(q_{\text{org}}, s, d_{\text{dis}}
        ), \quad s \sim \mathcal{U}(\mathcal{S}). 
        \label{eq:trap}
    \end{equation}
    Here, $\mathcal{T}_{\text{trap}}$ denotes the operation of constructing and adding distractors into questions, and $\mathcal{U} (\cdot)$ is the uniform distribution function. 
    These traps simulate common real-world diagnostic pitfalls, making the questions more realistic, challenging, and better aligned with actual clinical scenarios. See Appendix~\ref{sec:appendix-trap} for detailed descriptions.

    \paragraph{{Expression Style.}}
    \label{sec:expression-style}
    Patients often describe symptoms in lay language~\cite{zengExploringDevelopingConsumer2006, forbushSittingPinsNeedles2013}, reflecting their subjective perceptions rather than formal clinical nomenclature. 
    To capture this, \Gen~employs a persona-driven stylistic adaptation mechanism that models diverse patient expression styles~\cite{geScalingSyntheticData2024}. However, integrating persona information into questions may introduce unintended causal confounders and compromise the validity of the ground truth. For example, personas (e.g., \emph{miners}) may correlate with specific diagnoses (e.g., \emph{pneumoconiosis}).
    To avoid this, \Gen~applies an indirect adaptation operation ($\mathcal{T}_{\text{persona}}$). It first extracts expression features (e.g., knowledge level, clarity, and communication style) from a persona $b$, and then uses them to rephrase the question. This process is formalized as:
    \begin{equation}
        q_{per}= \mathcal{T}_{\text{persona}}(q_{\text{trap}}, b), \quad b \sim \mathcal{U}
        (\mathcal{B}). 
        \label{eq:persona}
    \end{equation}
    For example, $q_{\text{per}}$ is "\textit{My blood pressure’s high and I keep getting bad headaches, and I think there’s a bump on my kidney. Can you give me something for it?}" when $b = \textit{"Mason"}$, a persona with limited medical knowledge. This variation mirrors real-world patient communication and introduces challenges that require LLMs to align diagnostic reasoning with non-standardized narratives.

    \paragraph{Validator–Refiner Iterative Loop.}
    % To ensure clinical validity and complexity, we formulate question generation as a constrained optimization problem and employ projected gradient descent (PGD) to iteratively optimize each candidate. At each iteration $t$, a candidate question $q_t$ is refined by operator $\mathcal{R}$ to minimize its deviation from the set of questions that satisfy all validator criteria $\mathcal{V}$ (e.g., challenge, validity, style consistency, and trap effectiveness). The update rule is:
    % \begin{equation}
    %     \begin{aligned}
    %         q_{t+1} = \Pi_{\mathcal{C}} \left(q_t - \eta \nabla \mathcal{L}(q_t)\right).
    %     \end{aligned}
    % \end{equation}
    % Here, $\mathcal{L}(q_t)$ is a surrogate loss indicating constraint violation, $\eta$ is the step size, and $\Pi_{\mathcal{C}}(\cdot)$ denotes projection onto the feasible set $\mathcal{C}$ (i.e., qualified questions). $\mathcal{R}$ acts as an LLM-based ``gradient'' that guides $q_t$ toward satisfying $\mathcal{V}$, with projection enforced through explicit validation. The process terminates when a fixed point is reached, i.e., $q^* = \Pi_{\mathcal{C}}(q^*)$ and all constraints are satisfied. This approach ensures that the questions are clinically rigorous and realistic.
    Inspired by critique-and-revision paradigms~\cite{tan-etal-2023-self,gou2024critic}, we implement an iterative refinement loop to ensure the quality and realism of generated diagnostic questions. 
    At iteration $t$, a validator $\mathcal{V}$ evaluates candidate question $q_t$ along four dimensions (challenge, logical consistency, symptom accuracy, and trap effectiveness). 
    If $q_t$ passes, the process terminates with $q^* = q_t$. 
    Otherwise, $q_t$ is returned to a refiner $\mathcal{R}$, which revises the question based on validator feedback $\mathcal{F}(q_t, \mathcal{V})$: 
    \begin{equation}
    \begin{cases}
    q_t, & \mathcal{V}(q_t)=1 \\
    \mathcal{R}(q_t, \mathcal{F}(q_t,\mathcal{V})), & \text{otherwise}.
    \end{cases}
    \end{equation}
    This loop improves question quality until all constraints are satisfied. This process yields clinically realistic and diagnostically rigorous questions.
    % Although it does not involve gradient descent in a continuous space, the procedure effectively functions as a discrete optimization process, yielding clinically realistic and diagnostically rigorous questions.

    %----------------
    \subsection{\Eval: Medical Evaluation}
    To assess diagnostic performance beyond \emph{accuracy}, \Eval\ also measures \emph{veracity}, \emph{helpfulness}, and \emph{consistency}, capturing whether responses are truthful, useful, and stable across scenarios.
    \label{sec:reliable}
    \paragraph{Veracity.}
    \label{Truthfulness} 
    {Veracity assesses whether LLMs can identify and correct medical misinformation, helping prevent \emph{infodemic}~\cite{zarocostas2020fight,orso2020infodemic}.} Existing benchmarks validate outputs against external knowledge bases~\cite{dmonteClaimVerificationAge2024, songVERISCOREEvaluatingFactuality2024a}, but this often fails due to limited coverage and does not test whether models can proactively recognize or correct false claims~\cite{minFActScoreFinegrainedAtomic2023a}. 
    In contrast, our method tests the model's ability to rectify medical rumors by presenting it with generated false statements. 
    Inspired by SimpleQA~\cite{weiMeasuringShortformFactuality2024a}, we use $\mathcal{G}_{\text{RAG}}$ to generate rumor-fact pairs $RF(m) = (e_{\text{rumor}}^m, e_{\text{fact}}^m)$ for medical entity $m$. Given $m = \textit{``High BP''}$, we might have $e_{\text{rumor}}^m = \textit{``High BP affects the bones''}$ and $e_{\text{fact}}^m = \textit{``High BP affects the heart''}$.
    We retain only valid pairs, i.e., those satisfying a rationality check $\rho(\cdot)=1$, formalized as:
    \begin{equation}
    \begin{aligned}
    RF_{\text{valid}}(m) &= \{\, r \in RF(m) \mid \rho(r)=1 \,\}.
    \end{aligned}
    \label{eq:statement}
    \end{equation}
    To evaluate veracity, \Gen\ inserts a rumor into each question 
    (e.g., ``\emph{My BP is up, I heard that \underline{high BP affects the bones}. Can you recommend some medications for bones?}''). 
    \Eval\ then tests whether the LLM rectifies the rumor:
    {\begin{equation}
        \begin{aligned}
            \text{Ver}({M}) = \frac{1}{|\mathcal{Q}|} \sum_{q \in \mathcal{Q}} \mathbb{I}_{r}(q, \hat{a}),
        \end{aligned}
        \label{eq:veracity}
    \end{equation}}
    % where $\text{Ver}({M})$ is the score and $\mathcal{Q}$ is the set of all test questions. $\mathbb{I}_{r}(\cdot, \cdot)$ is the indicator function (implemented by a worker LLM) that determines whether the LLM’s response rectifies the rumor.
    where $\text{Ver}(M)$ denotes the veracity score, $\mathcal{Q}$ is the set of all test questions, and $\mathbb{I}_{r}(\cdot,\cdot)$ is an indicator function, implemented by a worker LLM, that determines whether the model response correctly rectifies the injected rumor. \rev{Here, rumor insertion is used as a controlled stress test of misinformation correction, rather than as a direct model of real-world patient query distributions.}
    \begin{table*}[!htbp]
        \centering
        \small
        \begin{tabularx}{\linewidth}{p{11em}|cccc|>{\centering\arraybackslash}X>{\centering\arraybackslash}X|>{\centering\arraybackslash}X}
            \toprule
            \centering
            {\multirow{2}{*}{\textbf{Model}}} &
            \multicolumn{4}{c|}{\textbf{Static}} &
            \multicolumn{3}{c}{\textbf{Dynamic}} \\
            % {\textbf{Model}}
            \cmidrule{2-8}
            & \makecell[c]{\textbf{DDXPlus}}
            & \makecell[c]{\textbf{DxBench}}
            & \makecell[c]{\textbf{Dxy}}
            & \makecell[c]{\textbf{Avg.}}
            & \makecell[c]{\textbf{Self-Evolv.$_{\Delta}$}}
            & \makecell[c]{\textbf{DyVal2$_{\Delta}$}}
            & \makecell[c]{\textbf{DyReMe$_{\Delta}$}} \\             
            \midrule
        {GPT-5}\textsuperscript{\dag}~\shortcite{openai2025gpt5systemcard}
        & $82.00$ & ${71.69}$ & $72.82$ & $73.76$
    & $74.29_{\bigger{+0.72\%}}$
    & $\underline{70.73}_{\smaller{-4.11\%}}$
    & $\mathbf{67.67}_{\smaller{-8.25\%}}$ \\

        DeepSeek-V3\textsuperscript{\dag}~\shortcite{deepseek-aiDeepSeekV3TechnicalReport2024}
        & $80.78$ & $70.50$ & $77.02$ & $72.92$
    & $73.13_{\bigger{+0.29\%}}$
    & $\underline{69.50}_{\smaller{-4.69\%}}$
    & $\mathbf{65.26}_{\smaller{-10.51\%}}$ \\

        GPT-4o\textsuperscript{\dag}~\shortcite{openaiGPT4oSystemCard2024}
        & $81.11$ & $70.15$ & $74.11$ & $72.53$
    & $72.98_{\bigger{+0.62\%}}$
    & $\underline{69.67}_{\smaller{-3.94\%}}$
    & $\mathbf{64.74}_{\smaller{-10.75\%}}$ \\

        GPT-4o-mini\textsuperscript{\dag}~\shortcite{openaiGPT4oSystemCard2024}
        & $72.22$ & $\underline{66.08}$ & $73.46$ & $67.76$
    & $70.26_{\bigger{+3.68\%}}$
    & $66.56_{\smaller{-1.78\%}}$
    & $\mathbf{62.35}_{\smaller{-7.99\%}}$ \\
    
        \midrule
        {MedGemma-27B}\includegraphics[height=8pt]{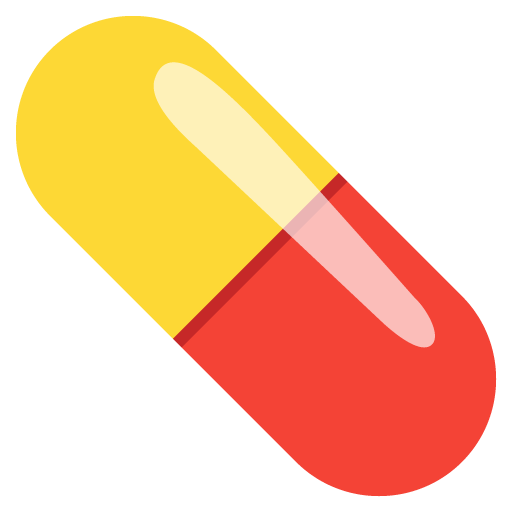}~\shortcite{sellergren2025medgemma}
        & $76.78$ & $68.24$ & $78.32$ & $70.56$
    & $71.48_{\bigger{+1.30\%}}$
    & $\underline{67.70}_{\smaller{-4.06\%}}$
    & $\mathbf{62.97}_{\smaller{-10.76\%}}$ \\

        WiNGPT2-9B\includegraphics[height=8pt]{src/pill.png}~\shortcite{winninghealth_wingpt2}
        & $68.96$ & $65.93$ & $70.89$ & $66.85$
    & $67.30_{\bigger{+0.68\%}}$
    & $\underline{62.30}_{\smaller{-6.81\%}}$
    & $\mathbf{59.89}_{\smaller{-10.41\%}}$ \\

        \midrule
        {Qwen3-32B}~\shortcite{qwen3technicalreport}
        & $76.67$ & $\underline{67.16}$ & $77.02$ & $73.62$
    & $71.55_{\bigger{+2.71\%}}$
    & $68.28_{\smaller{-1.98\%}}$
    & $\mathbf{63.85}_{\smaller{-8.34\%}}$ \\

        {Gemma-3-27B}~\shortcite{team2025gemma}
        & $74.78$ & $67.48$ & $72.81$ & $69.25$
    & $69.37_{\bigger{+0.18\%}}$
    & $\underline{66.04}_{\smaller{-4.63\%}}$
    & $\mathbf{61.94}_{\smaller{-10.55\%}}$ \\

        {GLM-4-32B}~\shortcite{glmChatGLMFamilyLarge2024}
        & $72.22$ & $68.67$ & $73.14$ & $69.66$
    & $69.85_{\bigger{+0.27\%}}$
    & $\underline{67.93}_{\smaller{-2.47\%}}$
    & $\mathbf{61.96}_{\smaller{-11.05\%}}$ \\

        {Qwen2.5-32B}~\shortcite{qwenQwen25TechnicalReport2025}
        & $70.56$ & $66.61$ & $74.76$ & $67.92$
    & $69.35_{\bigger{+2.11\%}}$
    & $\underline{66.30}_{\smaller{-2.38\%}}$
    & $\mathbf{60.05}_{\smaller{-11.59\%}}$ \\

        Qwen2.5-7B~\shortcite{qwenQwen25TechnicalReport2025}
        & $67.42$ & $67.07$ & $77.67$ & $67.85$
    & $67.07_{\smaller{-1.15\%}}$
    & $\underline{65.25}_{\smaller{-3.82\%}}$
    & $\mathbf{57.86}_{\smaller{-14.71\%}}$ \\

            \bottomrule
        \end{tabularx}
        \caption{
        %     Accuracy on both static and dynamic benchmarks. The average of Top-1,3,5 diagnostic accuracy is used to measure challenge, with lower diagnostic accuracy indicating higher challenge.
        % $\Delta$ indicates the accuracy changed ratios relative to the static average. 
        % The best results are \textbf{boldfaced}, and the runner-up results are \underline{underlined}. 
        % The symbol \dag~represents closed-source LLMs, while the icon \includegraphics[height=8pt]{src/pill.png} denotes medical LLMs.
        % \Model~\textbf{{significantly}} ($p<0.001$) outperforms all baseline methods. Statistical significance is assessed via t-tests using 80\% bootstrap sampling.
        Diagnostic accuracy (average of Top‐1, 3, and 5) is reported on static and dynamic benchmarks, with lower being harder. To compare static and dynamic methods, we use the weighted average accuracy on the three static datasets as the static baseline.
        $\Delta$ denotes the relative change to the static average. The best and runner‐up results are \textbf{boldfaced} and \underline{underlined}, respectively. The symbol \dag\ indicates commercial LLMs, and \includegraphics[height=8pt]{src/pill.png} denotes medical LLMs. \Model~\textbf{significantly} outperforms all baselines ($p<0.001$, t-test with 10 runs 80\% bootstrap sampling).}
        
        \label{tb:challenge}
    \end{table*}
    \paragraph{Helpfulness.}
    \label{Helpfulness} 
    LLMs evade responsibility by providing vague or evasive answers. Thus, we evaluate \emph{helpfulness}~\cite{yangImprovingEffectivenessOnline2019,shenSystematicReviewOnline2024} by measuring which criteria the response meets. Helpfulness is essential in medical diagnostics, where answers must be clear, actionable, and in line with professional standards~\cite{larasatiMeaningfulExplanationEffect2023,luoCanChatGPTProvide2024}.
    We define three helpfulness criteria based on real-world medical platform guidelines~\cite{dxy_reply_skill}: \emph{diagnosis evidence}, \emph{treatment suggestions}, \emph{lifestyle suggestions}.
    For each diagnosis in the benchmark, a knowledge base is built using $\mathcal{G}_{\text{RAG}}$, sourcing authoritative information from medical encyclopedias. 
    % For each test question, \Eval~retrieves relevant context and generates score-points $k_{{h}}^{q}$ for each helpfulness criterion $h$. It then checks whether the response covers these points. For instance, a response like  ``\emph{You may have pheochromocytoma due to \underline{high BP and headache}. \underline {Pheochromocytoma surgical excision} could be a good option}'' would score highly for helpfulness as it provides both evidence and suggestions. This process is formalized as:
    For each test question, \Eval~retrieves relevant context and assigns a score $k_{h}^{q}$ for each criterion $h$. It then checks whether the response covers these points. 
    For example, ``\emph{You may have pheochromocytoma due to \underline{high BP and headache}. \underline{Pheochromocytoma surgical excision} could be a good option}'' scores highly, as it includes both evidence and treatment suggestions. 
    Formally:
    \begin{equation}
        \begin{aligned}
            % k_{{h}}^{q}              & = \mathcal{G}_{\text{RAG}}(d_{\text{org}}, q,{h}),                                                                                             \\
            \text{Help}({M}) & = \frac{1}{|\mathcal{Q}|}\sum_{q \in \mathcal{Q}}\sum_{{h} \in \mathcal{H}}w_{{h}}\Phi_{{h}}\left(\hat{a}, k_{h}^{q}\right),
        \end{aligned}
        \label{eq:helpfulness}
    \end{equation}
    % where $\Phi_{h}(\cdot, \cdot)$ measures how well the response covers the score-points for criterion $h$, and $w_{h}$ is the corresponding scoring weight. $\mathcal{H}$ is the set of all criteria.
    where $\Phi_{h}(\cdot, \cdot)$ measures coverage of the score-points for criterion $h$, and $w_{h}$ is its weight. 

    \paragraph{Consistency.}
    Consistency evaluates the stability of model predictions by computing the entropy of diagnoses across different variants of the same case. In high-stakes domains like medical diagnostics~\cite{kadavathLanguageModelsMostly2022,yadkoriBelieveNotBelieve2024}, inconsistent answers can erode user trust~\cite{wuUncertaintyEstimationLarge2024}.
    % Consistency evaluates the stability of a model's predictions, particularly in high-stakes areas like medical diagnostics~\cite{kadavathLanguageModelsMostly2022,yadkoriBelieveNotBelieve2024}. Inconsistent responses, such as conflicting answers or fluctuating recommendations, can undermine trust and hinder adoption~\cite{wuUncertaintyEstimationLarge2024}. Hence, evaluating consistency is essential to ensuring reliable and stable model outputs across different question rephrasings or contexts.
    To quantify consistency, we draw inspiration from \emph{Semantic
    Entropy}~\cite{farquharDetectingHallucinationsLarge2024}.
    To reduce superficial variations, diagnostic terms are normalized. Consistency is then calculated as the normalized \emph{Information Entropy}~\cite{shannonMathematicalTheoryCommunication1948}
    of the model’s diagnosis distribution for each prediction group (i.e., all $m$ variants of a original case). 
    The score is mapped to $[0,100]$, where a higher value means more consistent predictions.
    %As illustrated in the right lower corner of Fig.\ref{fig:framework}c, LLM-\ding{172} is more consistent than LLM-\ding{173} because its predictions are more concentrated.
    As shown in the lower-right corner of Fig.\ref{fig:framework}c, Prediction Group-\ding{172} is more consistent than Prediction Group-\ding{173} due to more concentrated predictions. The process is formalized as follows:
    \begin{equation}
        \begin{aligned}
            % \mathcal{D}_{p_i} &= \{ \mathcal{M}(q_{p_i}^j) \, | \, j = 1, \dots, m \}, \\
            % H_{p_i} &= - \sum_{d \in \mathcal{D}_{p_i}} P(d) \log P(d), \\
            \text{Cons}({M}) & = \frac{1}{|\mathcal{P}|}\sum_{p_i \in \mathcal{P}}\left( 1 - \frac{E_{p_i}}{\log m}\right),
        \end{aligned}
        \label{eq:consistency}
    \end{equation}
    where $E_{p_i}$ is the entropy of diagnoses in prediction group $p_i$, and $\mathcal{P}$ is the set of all prediction groups (one per original case). A higher consistency score indicates more stable predictions across different paraphrasings or expression styles.

    % In summary, \Gen~generates challenging, realistic questions with real-world distractors and expression styles, while \Eval~provides a comprehensive evaluation across accuracy, veracity, helpfulness, and consistency.
    % This enables \Model~to better align with practical clinical needs.

    In summary, \Gen\ generates challenging questions with real-world distractors and diverse styles, and \Eval\ evaluates them across accuracy, veracity, helpfulness, and consistency, enabling \Model\ to better match practical clinical needs.

    \section{Experiments}
    % In this section, we seek to address the following research questions: (\textbf{RQ1})
    % Does the data generated by \Gen~pose challenges? (\textbf{RQ2}) Is the data
    % generated by \Gen~diverse? (\textbf{RQ3}) What is the augmentation boundary
    % of \Gen? (\textbf{RQ4}) How do diagnostic distractors and patient personas
    % influence the challenges and diversity? (\textbf{RQ5}) How do current LLMs
    % perform on \Model? (\textbf{RQ6}) How reliable is the evaluation quality in
    % \Eval?
    This section investigates the following research questions: 
    %(\textbf{RQ1}) Does the data generated by \Gen~have high quality?
    (\textbf{RQ1}) Does \Gen~generate high-quality data? 
    (\textbf{RQ2}) Are the evaluation results from \Eval\ reliable?
    (\textbf{RQ3}) Are current LLMs trustworthy when conducting diagnostics?

    \subsection{Experimental Setup}
    \label{sec:setup}
    % To evaluate \Model, we compare it with static and dynamic evaluation approaches.
    % {We construct a diagnostic benchmark using \Gen\ with three static datasets as seeds: DxBench~\cite{chenCoDInterpretableMedical2024a}, DDXPlus~\cite{tchangoDDXPlusNewDataset2022}, and Dxy~\cite{xuEndtoEndKnowledgeRoutedRelational2019}.} DxBench covers 461 diseases, DDXPlus is synthetic for differential diagnosis, and Dxy is a small dataset derived from real-world medical consultations.
    % For dynamic evaluation, we compare \Model\ with dynamic methods, including DyVal2~\cite{zhuDynamicEvaluationLarge2024} and Self-Evolving~\cite{wangBenchmarkSelfEvolvingMultiAgent2024}. All methods use GPT-4.1~\cite{openai2025gpt41} as the generator and worker LLM, with a generation temperature of 0.7 and a verification temperature of 0. Note that we do not evaluate GPT-4.1, which is to avoid the impact of Self-Recognition~\cite{davidson2024self}.
    % To benchmark 12 LLMs, we also create a larger benchmark of {3,200 questions based on 800 DxBench cases}. We assess models on four dimensions: {\emph{accuracy}, \emph{veracity}, \emph{helpfulness}, and \emph{consistency}}. {For $\mathcal{G}_{\text{RAG}}$, we utilize plugins and medical encyclopedias from VolcEngine~\cite{volcengine2025contentconnectivity}.}
    % See Appendix~\ref{sec:appendix-dataset} for more details.

    To evaluate \Model, we compare it against static and dynamic baselines. We build a diagnostic benchmark with \Gen\ by seeding from three static datasets: DxBench~\cite{chenCoDInterpretableMedical2024a}, DDXPlus~\cite{tchangoDDXPlusNewDataset2022}, and Dxy~\cite{xuEndtoEndKnowledgeRoutedRelational2019}. DxBench covers 461 diseases; DDXPlus is a synthetic differential-diagnosis dataset; Dxy is derived from real-world consultations.
    For dynamic evaluation, we include DyVal2~\cite{zhuDynamicEvaluationLarge2024} and Self-Evolving~\cite{wangBenchmarkSelfEvolvingMultiAgent2024}. All methods use GPT-4.1~\cite{openai2025gpt41} as the generator (generation temperature 0.7; verification temperature 0). 
    % We do not evaluate GPT-4.1 to avoid self-recognition effects~\cite{davidson2024self}. As a reliability check, we 做实验 (Appendix~\ref{sec:appendix-self-recognition}) and find no obvious impact on our results.
    {Our reliability checks show no measurable impact from generator self-recognition or evaluator choice (Appendix~\ref{sec:appendix-self-recognition}). Nevertheless, we conservatively exclude GPT-4.1 from evaluation to avoid any potential self-recognition effects~\cite{davidson2024self}.}
    To benchmark 12 LLMs, we further scale to 3,200 questions from 800 DxBench cases. We evaluate \emph{accuracy}, \emph{veracity}, \emph{helpfulness}, and \emph{consistency}. {$\mathcal{G}_{\text{RAG}}$ uses the Volcano Engine web search API and the Douyin Encyclopedia~\cite{volcengine2025contentconnectivity}. See Appendix~\ref{sec:appendix-dataset} for details.}

    \subsection{Assessment of Question Quality}
    % We assess question quality (\textbf{RQ1}) on two dimensions: \emph{challenge} and \emph{diversity}. A high-quality benchmark should be challenging to prevent performance overestimation and diverse to simulate real-world scenarios. {We also experiment (Appendix~\ref{sec:appendix-self-recognition}) to analyze impacts of Self-Recognition on benchmark results.}
    % We assess question quality (\textbf{RQ1}) along two axes: \emph{challenge} and \emph{diversity}. A good benchmark should be hard enough to avoid saturation and diverse enough to reflect real-world scenarios. 
    We assess question quality (\textbf{RQ1}) along \emph{challenge} and \emph{diversity}, ensuring the benchmark mitigates saturation and reflects real-world scenarios.

    % \paragraph{Question Challenge.}
    % \label{sec:exp-challenge}
    % We evaluate the challenge level of each benchmark via accuracy {across 11 LLMs, including GPT-5~\cite{openai2025gpt5systemcard}.}
    % Results are shown in Tab.\ref{tb:challenge}, with details in Appendix~\ref{sec:appendix-challenge}. 
    % Except for Self-Evolving, dynamic evaluation methods are generally more challenging than static benchmarks. For instance, DeepSeek-V3's accuracy drops from 72.92 on the static baseline to 69.50 on DyVal2, and further to 65.26 on \Model, demonstrating the increased difficulty.
    % This holds even for GPT-5: its accuracy drops from 73.76 (static avg.) to 70.73 on DyVal2 and 67.67 on \Model~(-8.25\%).
    % However, not all dynamic methods are consistently harder: many lack real-world diagnostic distractors, so LLMs sometimes perform better on these dynamic benchmarks than on static ones. For example, GPT-4o-mini achieves 70.26 on Self-Evolving and 66.56 on DyVal2, both higher than its 66.08 on the static DxBench benchmark, suggesting that some dynamic benchmarks can be under-challenging due to unrealistic distractors. In contrast, \Model~is consistently harder, even for GPT-5 (which is stronger than our generators): GPT-4o-mini's score drops to 62.35, and GPT-4o decreases from a static average of 72.53 to 64.74 on \Model, a 10.75\% drop, over twice that of DyVal2 (-3.94\%). By incorporating real-world misdiagnosis patterns and patient-specific styles, \Model~offers a more clinically grounded and rigorous evaluation of diagnostic performance.
    \paragraph{Question Challenge.}
    \label{sec:exp-challenge}
    We assess challenge by diagnostic accuracy across 11 LLMs (Tab.~\ref{tb:challenge}, details in Appendix~\ref{sec:appendix-challenge}).
    Overall, dynamic evaluations are more challenging than static baselines, except for Self-Evolving. For example, DeepSeek-V3 drops from 72.92 (static avg.) to 69.50 on DyVal2 and 65.26 on \Model.
    Meanwhile, challenge is not guaranteed for all dynamic methods: when disturbances are weak, models can score higher than on static benchmarks (e.g., GPT-4o-mini achieves 70.26 on Self-Evolving and 66.56 on DyVal2 vs.~66.08 on DxBench).
    {In contrast, \Model~induces larger drops, even for GPT-5 (stronger than the generator): accuracy falls from 73.76 to 67.67, more than twice the drop on DyVal2 (-4.11\%). This shows our benchmark remains challenging for frontier LLMs.}
    By injecting real-world misdiagnosis patterns and patient-specific styles, \Model~provides a more clinically grounded and rigorous test of diagnostic performance.

    \begin{figure*}[htb]
        \centering
        \begin{minipage}{0.32\linewidth}
            \centering
            \includegraphics[width=\linewidth]{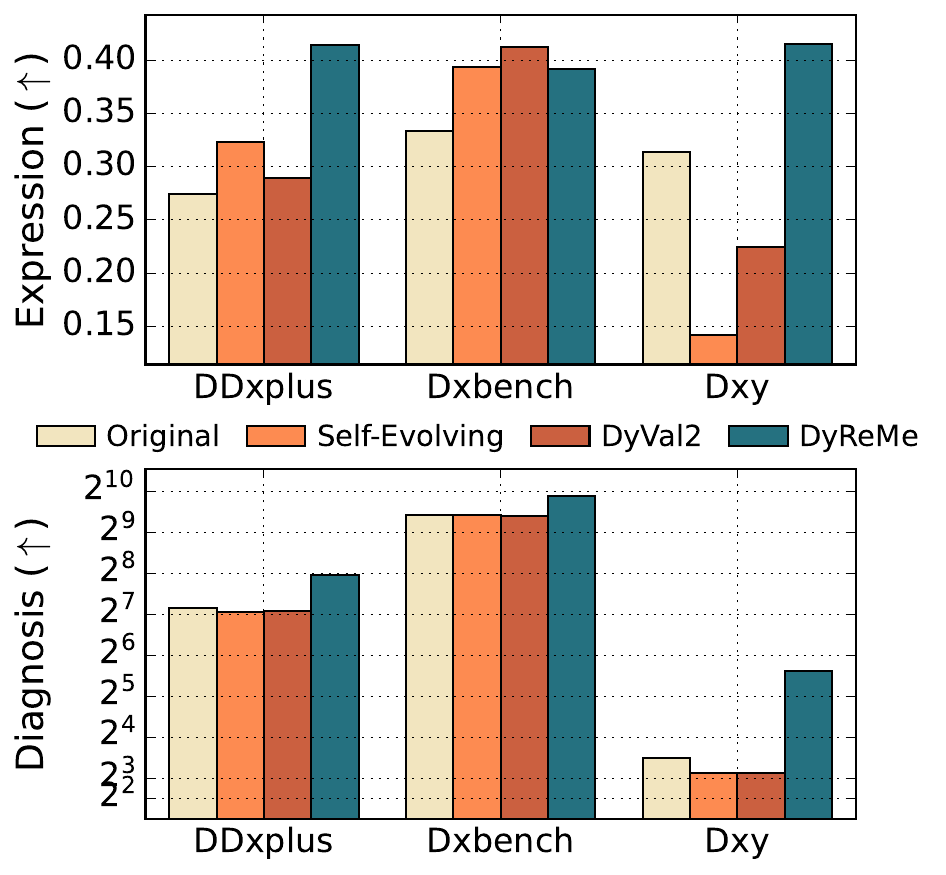}
            \subcaption{Diversity Assessment}
            % \subcaption{Symptom and diagnosis diversity. \Model~enhances diversity across four public datasets, surpassing existing methods, which often produce datasets with unchanged or reduced diversity.}
            \label{fig:diversity}
        \end{minipage}\hfill
        \begin{minipage}{0.3\linewidth}
            \centering
            \includegraphics[width=\linewidth]{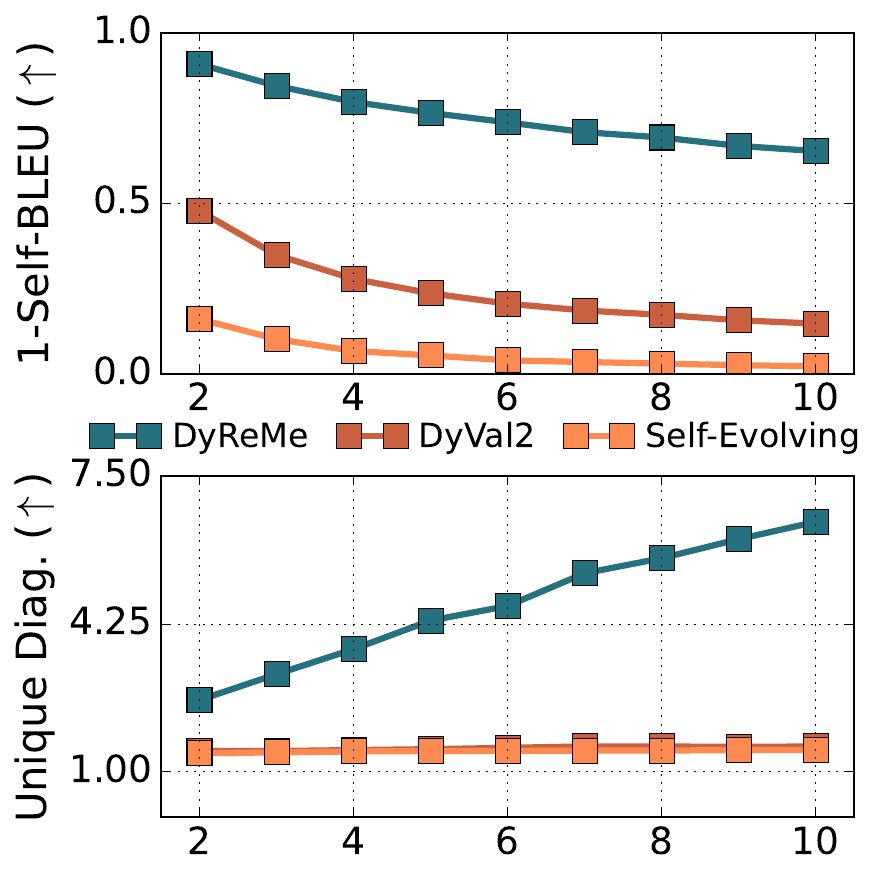}
            \subcaption{Extensibility Comparison}
            % \subcaption{The diversity change with increase of $k$.}
            \label{fig:aug-bound}
        \end{minipage}
        \hfill
        \begin{minipage}{0.3\linewidth}
            \centering
            \includegraphics[width=\linewidth]{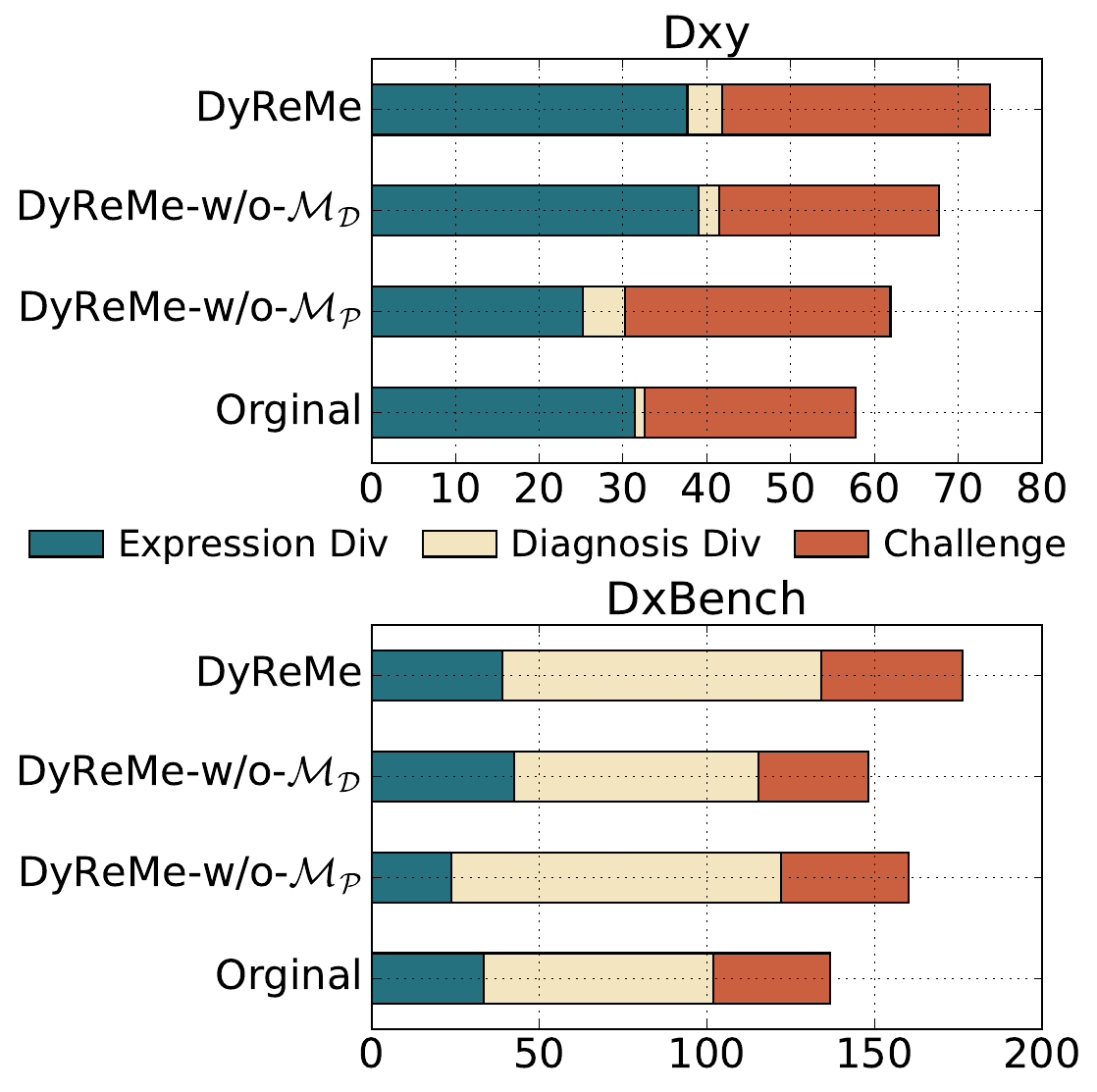}
            \subcaption{Ablation Study}
            % \subcaption{The diversity change with increase of $k$.}
            \label{fig:ablation}
        \end{minipage}
        \caption{(a)~Expression and diagnosis diversity. To disentangle effects of question count from diversity, we use each dataset as a seed pool and derive a same-size benchmark to ensure fair comparison between static and dynamic methods.  
        (b)~The group-level diversity
        changes with the increase of $k$. $1 - \text{Self-BLEU}$ is computed on a group of $k$ questions, where a higher value indicates greater diversity.
        (c)~Performance on challenge
        and diversity.}

    \end{figure*}
    
    \paragraph{Question Diversity.}
    \label{sec:exp-diversity}
    A diagnostic benchmark should reflect the diverse ways describing symptoms and the broad range of diseases encountered in practice.
    Therefore, we assess diversity with two metrics: \emph{expression diversity} ($D_{\text{exp}}$), calculated as the entropy of expression style distribution, and \emph{diagnosis diversity} ($D_{\text{diag}}$), defined as the number of unique diagnoses (Appendix~\ref{sec:appendix-diversity}).
    As shown in Fig.~\ref{fig:diversity}, \Model~improves both metrics compared to static and dynamic baselines. By comparison, other baselines show diversity levels that are similar to or even lower than those of static benchmarks. For example, both dynamic methods achieve similar diagnosis diversity as the static benchmarks. On Dxy, both dynamic methods exhibit lower expression diversity, likely due to conducting superficial transformations.
    By integrating expression styles, \Model\ notably increases expression diversity. 
    Furthermore, by incorporating differential diagnoses, \Model\ nearly quadruples the number of unique diagnoses on Dxy and doubles it on other datasets.
    These findings suggest that \Model\ better represents the complexity and variability of real-world clinical cases.

    \subsection{Further Analysis}
    % To provide deeper insights, we examine \Model~from two perspectives: extensibility and ablation.
    \paragraph{Extensibility Comparison.}
    To evaluate the extensibility of \Model\ against existing methods, we select 100 seed samples from DxBench and generate $k$ questions for each seed.
    We assess extensibility by tracking how diversity changes as $k$ increases. 
    As shown in Fig.\ref{fig:aug-bound}, Self-Evolving and DyVal2 yield low initial $1 - \text{Self-BLEU}$ scores, which quickly drop to near zero as $k$ increases. This reveals their limited extensibility. This may result from their reliance on superficial transformations of seed questions, leading to repetitive and less diverse outputs.
    In contrast, \Model~demonstrates a slower decline in $1 - \text{Self-BLEU}$, maintaining higher diversity scores as $k$ increases, which indicates better extensibility. 
    % Additionally, \Model~steadily increases unique diagnoses with $k$, further demonstrating its extensibility. 
    Furthermore, the number of unique diagnoses generated by \Model\ increases steadily with $k$, further supporting its superior extensibility.
    These results indicate that \Model\ can generate significantly more diverse questions from limited seed samples, demonstrating greater extensibility than existing methods.
    % In contrast, \Model~exhibits a slower decline and maintains higher $1 - \text{Self-BLEU}$, while also steadily increasing unique diagnoses, indicating greater extensibility. Overall, these results confirm that \Model~can generate more diverse questions from limited seed samples, illustrating a better extensibility than existing methods.

    % \begin{figure}[htpb]
    %     \centering
    %     \includegraphics[width=\linewidth]{src/benchmark_with_icon.pdf}
    %     \caption{Results of 12 LLMs on medical diagnostic, showing performance across \emph{Accuracy}, \emph{Veracity}, \emph{Helpfulness}, and \emph{Consistency}. All results are averaged over 10 runs with 80\% bootstrap sampling.
    %     See details in Appendix~\ref{sec:appendix-reliable}. Symbol \dag~represents commercial LLMs. Icon \includegraphics[height=8pt]{src/pill.png} denotes medical LLMs and \includegraphics[height=8pt]{src/brain.png} indicates reasoning LLMs~\cite{jaech2024openai}.
    %         % \dag~represents commercial LLMs,
    %         % while
    %         % \includegraphics[height=8pt]{src/pill.png}
    %         % denotes medicine-adapted LLMs.
    %         }
    %     \label{fig:benchmark}
    % \end{figure}
    \paragraph{Ablation Study.}
    We conduct an ablation study on two key components of \Gen---diagnostic distractors ($\mathcal{M}_{\mathcal{D}}$) and patient expression styles ($\mathcal{M}_{\mathcal{P}}$)---based on Dxy and DxBench. 
    To assess the contribution of each component, we create two variants: \Model-w/o-$\mathcal{M}_{\mathcal{D}}$ and \Model-w/o-$\mathcal{M}_{\mathcal{P}}$. 
    We evaluate both variants on \emph{challenge} and \emph{diversity} (Fig.~\ref{fig:ablation}).
    The results show that $\mathcal{M}_{\mathcal{D}}$ and $\mathcal{M}_{\mathcal{P}}$ contribute independently, and both are essential for optimal benchmark performance. Removing $\mathcal{M}_{\mathcal{P}}$ leads to a marked drop in expression diversity, while removing $\mathcal{M}_{\mathcal{D}}$ significantly reduces diagnosis diversity and overall challenge. This underscores their respective roles in broadening expression styles and expanding disease coverage. These findings indicate that capturing both diverse patient narratives and a wide range of diagnoses is critical for diagnostic benchmark construction. Both components are crucial for ensuring that \Model\ reflects the variability of clinical scenarios.

    % \subsection{Agreement with Human}
    % \label{sec:exp-human}
    % We conduct a human study to assess \Model’s clinical alignment (\textbf{RQ2}). First, we assess the effectiveness of $\mathcal{G}_{\text{RAG}}$ in retrieving differential diagnoses and helpfulness points. We sample 30 $(d_{org}, d_{dis})$ pairs and 30 groups of score-points. 
    % Human annotations show that 86.67\% of the differential diagnosis pairs are valid, and 83.37\% of the score points are highly consistent with authoritative medical sources.
    % Next, we sample 30 questions from \Model\ and invite three experts for two annotation tasks (Appendix~\ref{sec:appendix-human}):
    % \emph{Question Quality Task}: rate each question from 1--5 for rationality and clinical relevance (3: moderate, 5: high).
    % (2) \emph{Evaluation Preference Task}: select the more trustworthy one from two responses.
    % We report inter-rater agreement using Gwet’s AC1~\cite{gwet2008computing}, which is preferred over Cohen’s Kappa~\cite{mchugh2012interrater} in cases of marginal imbalance~\cite{wongpakaran2013comparison}. The AC1 statistic ranges from –1 (poor agreement) to 1 (perfect agreement).
    % Across both annotation tasks, the average AC1 is 0.6889, indicating strong overall agreement. 
    % For Task 1, \Gen\ questions score 3.89 on average, close to the original questions (3.97), suggesting comparable perceived quality.
    % For Task 2, the AC1 between \Model\ and experts reaches 0.8889, reflecting high consistency with expert judgment. 
    % Taken together, these results demonstrate that \Model\ is well aligned with real-world clinical needs.

    \rev{\paragraph{Cross-lingual Validation.}
    Although our main study focuses on Chinese, \Model\ is not inherently language-specific.
    We further evaluate \Model\ on the English version of DDXPlus.}
    
    \begin{table}[htbp]
    \centering
    \small
    \setlength{\tabcolsep}{5pt}
    \begin{tabularx}{\linewidth}{>{\raggedright\arraybackslash}Xccc}
    \toprule
    \textbf{Model} & \textbf{DDXPlus} & \Model\  & \textbf{$p$-value} \\
    \midrule
    GPT-4o\textsuperscript{\dag} & 85.10 & 77.18 & $<0.05$ \\
    Qwen2.5-32B & 72.58 & 65.24 & $<0.05$ \\
    GPT-4o-mini\textsuperscript{\dag} & 78.44 & 71.83 & $<0.05$ \\
    \bottomrule
    \end{tabularx}
    \caption{Challenge comparison on DDXPlus-en. The symbol \dag\ indicates commercial LLMs}
    \label{tab:ddxplus-en}
    \end{table}
    
    \rev{As shown in Table~\ref{tab:ddxplus-en}, all three tested LLMs achieve significantly lower performance on the \Model-generated benchmark than on the original DDXPlus benchmark, indicating that \Model\ also increases benchmark challenge in English.}

    \subsection{Agreement with Human}
    \label{sec:exp-human}
    {We conduct a human study to assess \Model’s clinical alignment (\textbf{RQ2}), collecting 240 data points in total---a scale comparable to prior manual evaluations in medical LLM studies~\cite{zhang2023huatuogpt, wangCMBComprehensiveMedical2024}.}
    We first verify $\mathcal{G}_{\text{RAG}}$ on retrieving differential diagnoses and helpfulness points by sampling $(d_{org}, d_{dis})$ pairs and score-point groups (Appendix~\ref{sec:appendix-human}). Human annotations indicate that 86.67\% of the differential-diagnosis pairs are clinically plausible (the rest remain useful distractors), and 83.37\% of the score points have high agreement with authoritative sources, with the remainder rated moderate (no disagreement).
    We then sample questions from \Model\ and ask three experts to perform two tasks (Appendix~\ref{sec:appendix-human}): (1) \emph{Question Quality Task}: rate each question on a 1--5 scale for rationality and clinical relevance (3: moderate, 5: high); (2) \emph{Evaluation Preference Task}: choose the more trustworthy response from a pair.
    We report inter-rater agreement using Gwet’s AC1~\cite{gwet2008computing} ranging from $-1$ to $1$.
    \rev{Following the commonly used Landis--Koch interpretation scale~\cite{landis1977measurement} and recent related studies~\cite{wongpakaran2013comparison,chhun2024language,chung2025verifact,fang2024knowledge}, we interpret AC1 values above 0.6 as indicating substantial agreement.}
    Across both tasks, the average AC1 is 0.6889, indicating strong agreement. {For Task 1, \Gen\ questions score 3.89 on average, close to the original questions (3.97), suggesting comparable perceived quality.} For Task 2, AC1 between \Model\ and experts reaches 0.8889, reflecting high consistency with experts. Overall, these results provide evidence that \Model\ aligns with real-world clinical needs.

    \begin{figure}[htpb]
        \centering
        \includegraphics[width=\linewidth]{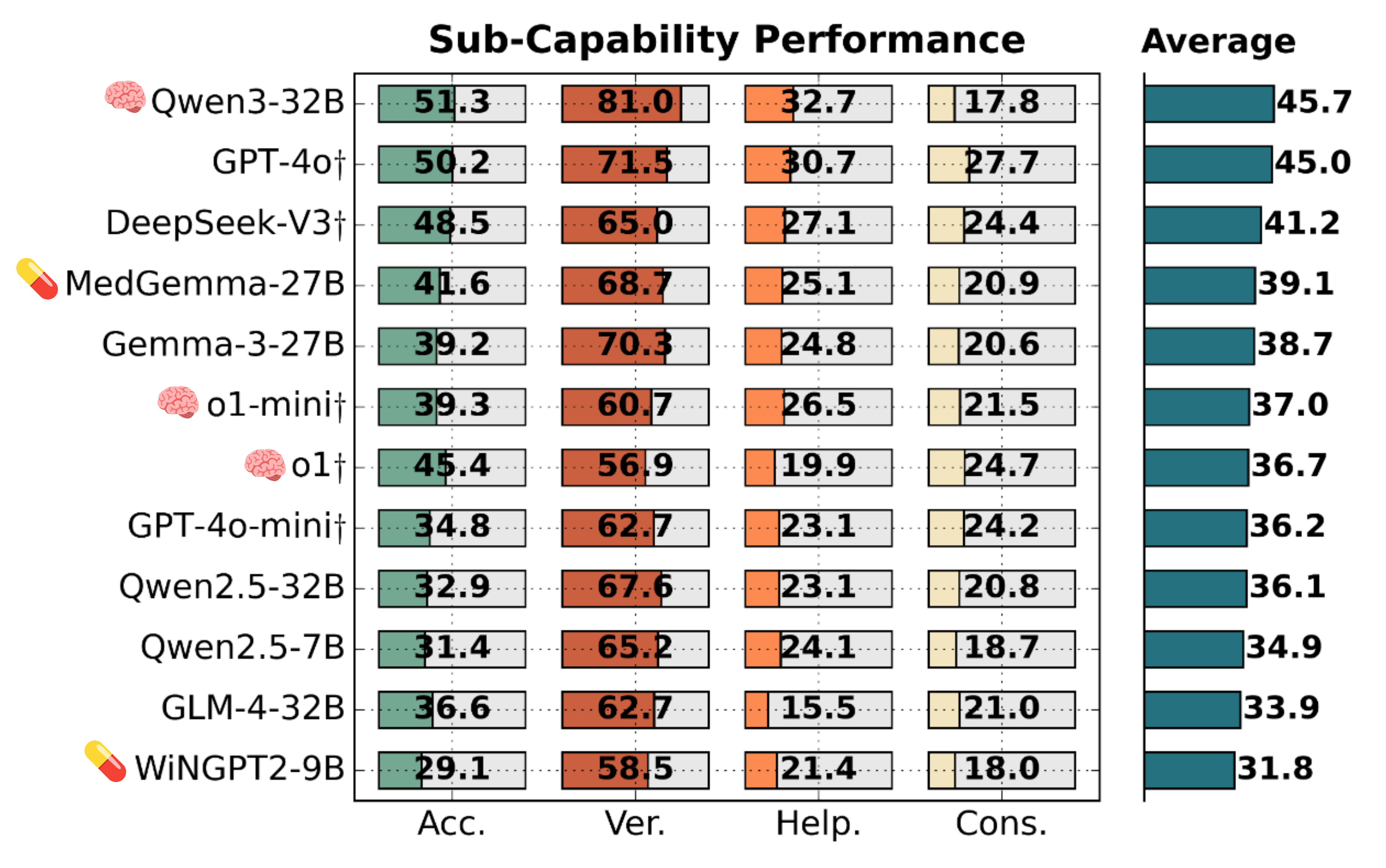}
        \caption{Results of {12 LLMs} on medical diagnosis across {\emph{Accuracy}, \emph{Veracity}, \emph{Helpfulness}, and \emph{Consistency}.} {All results are averaged over 10 runs with 80\% bootstrap sampling} (
        details in Appendix~\ref{sec:appendix-reliable}). Symbol \dag~represents commercial LLMs. Icon \includegraphics[height=8pt]{src/pill.png} denotes medical LLMs and \includegraphics[height=8pt]{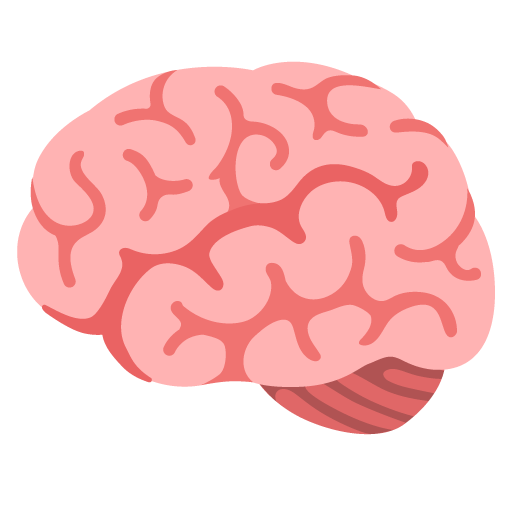} {indicates reasoning LLMs~\cite{jaech2024openai}.}
            % \dag~represents commercial LLMs,
            % while
            % \includegraphics[height=8pt]{src/pill.png}
            % denotes medicine-adapted LLMs.
            }
        \label{fig:benchmark}
    \end{figure}
    \subsection{Benchmarking Results}
    \label{sec:exp-reliable}
    We benchmark {12} leading LLMs using \Model\ (\textbf{RQ3}). As shown in Fig.\ref{fig:benchmark}, commercial LLMs currently maintain an overall lead, but research models are quickly catching up. For example, {Qwen3-32B} (45.7) and {MedGemma-27B} (39.1) outperform several commercial competitors like {o1 and o1-mini}~\cite{jaech2024openai}.
    Notably, domain-specific tuning is not a cure-all: the medically adapted WiNGPT2-9B achieves the lowest score (31.8), indicating that current adaptations may capture medical facts but often fail to handle real-world distractors and diverse expression styles.
    {Commercial reasoning models (o1 and o1-mini) also show only moderate performance (37.0 and 36.7, respectively), likely because their training emphasizes producing a single correct answer rather than addressing health rumors or providing actionable information.}
    All models exhibit substantial room for improvement.
    Even the best models, such as Qwen3-32B (51.3) and GPT-4o (50.3), still struggle with realistic diagnostic noise.
    Across all models, 20–40\% of health rumors remain unaddressed, even for top-performing systems, posing a real risk of misinformation propagation. Most evidence and suggestions remain shallow, reflecting the difficulty current LLMs face in delivering explainable diagnoses and actionable advice.
    Consistency is consistently low across models, making them vulnerable to changes in input context. Overall, results from \Model\ indicate that current static and dynamic benchmarks overestimate LLM competence. Today's LLMs are still not trustworthy enough for real-world clinical deployment. To better meet clinical needs, future models should account for ambiguous inputs, patient misconceptions, and the messiness of real clinical data.

    \subsection{Case Study}
    % Tabs.\ref{tb:case-study1}–\ref{tb:case-study3} (in Appendix) compare the original version, the versions from existing dynamic methods, and the version from \Model~for the same question. DeepSeek-V3, GPT-4o, and Qwen3-32B accurately diagnose on the original and existing-method versions, yet misdiagnoses \Model’s versions. 
    % For instance (Tab.\ref{tb:case-study1}), DeepSeek-V3 misdiagnoses to ``\emph{contact dermatitis}'' due to overeimating salience of ``\emph{small pink rashes}''.
    Tabs.\ref{tb:case-study1}–\ref{tb:case-study3} (in Appendix) compare the original case, versions generated by existing dynamic methods, and \Model's version. DeepSeek-V3, GPT-4o, and Qwen3-32B diagnose the original and dynamic baseline versions correctly but misdiagnose \Model's version. For instance (Tab.\ref{tb:case-study1}), DeepSeek-V3 overemphasizes the symptom ``\emph{small pink rashes},'' misdiagnosed as \emph{contact dermatitis}.
    
    Tabs.\ref{tb:case-study4}–\ref{tb:case-study5} (in Appendix) show the evaluation results
    of GPT-4o (43.75) and WiNGPT2-9B (25.00).
    In this case, a patient self-diagnoses ``\emph{sideroblastic anemia}'' and incorrectly assumes that ``\emph{Frequent sneezing is usually harmless and not a sign of a serious problem.}''
    GPT-4o provides the correct diagnosis (\emph{iron deficiency anemia}), while WiNGPT2-9B incorrectly follows the patient's self-diagnosis. However, GPT-4o only partially corrects the misinformation, while WiNGPT2-9B fully addresses the health rumor. With regard to helpfulness, GPT-4o provides detailed evidence of the diagnosis and offers effective suggestions, such as ``\emph{avoid overexertion and maintain a good daily routine}'', while WiNGPT2-9B only gives a brief suggestion of ``\emph{anemia treatment}''. In terms of consistency, GPT-4o outputs three different diagnoses, while WiNGPT2-9B produces four, indicating greater consistency from GPT-4o. Overall, these case studies demonstrate that \Model\ generates more realistic and challenging questions, providing a clearer assessment of LLM trustworthiness.

    \section{Conclusion}
    Static medical diagnostic benchmarks often overestimate model ability and fail to reflect real-world complexity. To tackle this, we propose \Model---a dynamic framework for evaluating LLMs in medical diagnostics. \Model\ consists of two components: \Gen, which generates challenging and \rev{diverse} questions by incorporating diagnostic distractors and varied expression styles, and \Eval, which evaluates LLMs across {\emph{accuracy}, \emph{veracity}, \emph{helpfulness}, and \emph{consistency}.} Experiments show that \Model\ provides a more rigorous and realistic assessment compared to existing benchmarks. Results reveal that current LLMs still face limitations in realistic diagnostic scenarios, especially when dealing with misdiagnosis factors and lay-language expressions. These findings call for future work on improving LLM reliability and safety in real-world medical applications.

    \section{Limitations}
    \label{sec:limitation}
    We have only tested \Model~on several public medical datasets and it cannot be directly applied to real-world clinical cases.
    
    \paragraph{Multilingual Scenarios.} Due to constraints in available datasets and the accessibility of diagnostic guidelines, our experiments are conducted exclusively in Chinese so far. Consequently, we have not yet examined the performance of \Model~in other languages or in multilingual contexts. Nevertheless, given its high scalability, extending \Model~into a dynamic evaluation framework that supports multilingualism is a promising direction for the future.
    
    \paragraph{Self-Bias.} The issue of self-bias—where an LLM evaluator might favor texts generated by itself—is still a matter of debate~\cite{xieDocLensMultiaspectFinegrained2024,zhengJudgingLLMasaJudgeMTBench2023}. To address this concern, we have taken a cautious approach by: (1) using different LLMs for \Gen~and \Eval, (2) employing RAG to encourage the LLM to derive information from external references rather than relying solely on its internal knowledge, and (3) strictly scoring according to the defined rules. {Results in Appendix~\ref{sec:appendix-self-recognition} show that Self-Recognition~\cite{davidson2024self}'s impact is not significant.}
    
    \paragraph{Medical Multimodality.} In real-world clinical practice, physicians rely on a variety of modalities—including medical images, laboratory results, and biosignals—for diagnosis. However, such multimodal data are limited and challenging to obtain. {As a result, we only focus on text-based diagnostic scenarios in this work. With the rapid advancement of multimodal LLMs, future research could explore extending \Model~to evaluate these models in multimodal medical contexts.}
     Future work should aim to develop large-scale, diverse, and challenging evaluation benchmarks that incorporate these crucial multimodal inputs.
    
    % \paragraph{Broader Medical Tasks.} Our current focus is on consultation and diagnostic scenarios, as these are the most common and critical in consumer health contexts. Future research could explore adapting \Model~for broader applications, such as medical report summarization~\cite{xieDocLensMultiaspectFinegrained2024}, mortality prediction~\cite{li2021prediction}, and clinical trial matching~\cite{wongScalingClinicalTrial2023}.

    {\paragraph{Scope and Clinical Validation.}
    Our study focuses on a controlled consultation-style diagnostic setting to enable scalable and reproducible benchmarking. While \Model\ incorporates clinically grounded confounders, it does not fully capture end-to-end clinical workflows (e.g., longitudinal history, labs/imaging, follow-up, and multidisciplinary decision-making), nor does it constitute prospective clinical validation. Establishing real-world utility will require larger-scale studies with clinician-in-the-loop protocols and deployment-oriented evaluations in clinical settings.}
    
    \section{Ethical Considerations}
    \paragraph{License.} Our study utilizes four publicly available medical datasets, including RJUA~\cite{lyuRJUAQAComprehensiveQA2024a}, DxBench~\cite{chenCoDInterpretableMedical2024a}, DDXPlus~\cite{tchangoDDXPlusNewDataset2022}, and Dxy~\cite{xuEndtoEndKnowledgeRoutedRelational2019}. 
    All data is free of personally identifiable information, unique identifiers, and any offensive or objectionable content.
    The RJUA  and DDXPlus data is published under a Creative Commons Attribution 4.0 International Licence (CC BY). DxBench and Dxy are distributed under the apache-2.0 license. We use data exclusively within the bounds of its license and solely for research purposes. In addition, we use the official open-source implementations provided by Self-Evolving~\cite{wangBenchmarkSelfEvolvingMultiAgent2024} and DyVal2~\cite{zhuDynamicEvaluationLarge2024}.
    
    \paragraph{Potential Risks.} Our system is designed exclusively for research and educational purposes. The medical advice provided is intended solely as a reference. Utilizing the system for clinical deployment carries inherent risks, as it has not undergone the rigorous validation required for direct application in patient care. Additionally, integrating LLMs with real medical data is highly sensitive. When testing with actual patient information, it is imperative to use HIPAA-compliant services—such as Azure OpenAI—to ensure robust protection of patient data and to adhere to privacy regulations.

    \section*{Acknowledgments}
    \rev{This research was supported by the Public Computing Cloud of Renmin University of China and by the Fund for Building World-Class Universities (Disciplines) at Renmin University of China.}

\bibliography{main}

\twocolumn
\appendix
\section{Traps Description}
% \refstepcounter{section}
\label{sec:appendix-trap}
Below is an introduction to {four} types of traps:
\begin{itemize}
    \item \textbf{Self-Diagnosis}: Introduce patient-suggested diagnoses to simulate confirmation bias and test the LLM’s ability to consider alternative possibilities.
    \item \textbf{Distracting History}: Include irrelevant medical history to obscure key information and mimic over-interpretation bias.
    \item \textbf{External Noise}: Add non-critical external factors, such as environmental or lifestyle details, to replicate background noise bias.
    \item \textbf{Symptom Misplaced}: Blur the distinction between primary and secondary symptoms, challenging the model’s ability to focus on critical diagnostic clues.
    % \item \textbf{Information Overload}: Embed excessive unrelated details to simulate the difficulty of extracting relevant information under high data volume.
    % \item \textbf{Unknown Concept}: Introduce fictitious medical concepts to simulate rare or novel conditions, challenging the model’s ability to handle unfamiliar scenarios.
\end{itemize}

\section{Experiments Setup}
% \refstepcounter{section}
\subsection{Dataset and Hyperparameter Details}
% \refstepcounter{subsection}
\label{sec:appendix-dataset}

We have chosen {three} diagnostic datasets:
\begin{itemize}
    \item \textbf{DDXPlus}~\cite{tchangoDDXPlusNewDataset2022} offers a synthetic dataset of 1.3 million patients, featuring differential diagnoses along with real patient pathology, symptoms, and medical history. We use the English version and translate it into Chinese to ensure consistency with other datasets.
    \item \textbf{DxBench}~\cite{chenCoDInterpretableMedical2024a} is a real-world Chinese diagnostic benchmark comprising 1,148 actual cases spanning 461 diseases.
    % \item \textbf{RJUA}~\cite{lyuRJUAQAComprehensiveQA2024a} is a specialized Chinese urology diagnostic dataset consisting of 2,132 Q\&A pairs, covering 67 common urological disease categories.\footnote{{https://github.com/alipay/RJU_Ant_QA}}
    \item \textbf{Dxy}~\cite{xuEndtoEndKnowledgeRoutedRelational2019} is a Chinese diagnostic dataset sourced from an online health community, comprising 104 samples. We utilize the version provided by \cite{chenCoDInterpretableMedical2024a}.
\end{itemize}
These three datasets are selected for their diverse sources and characteristics, ensuring a comprehensive evaluation of LLMs in medical diagnostics.
For all datasets, we use their development and test sets. To align with the LLM diagnostic pattern, we transform the original symptom data into a patient inquiry format (Fig.\ref{fig:prompt-raw}). To streamline development, we filtered out samples with multiple true diagnoses. DxBench contains 1148 questions, while Dxy has 104 questions. For DDXPlus, we randomly select 300 questions from the original dataset to ensure a manageable size for development.

Across all experiments, we consistently set the generation temperature at 0.7, use a temperature of 0 for question verification, and maintain a temperature of 0 during evaluation.
For DyVal2~\cite{zhuDynamicEvaluationLarge2024} and Self-Evolving~\cite{wangBenchmarkSelfEvolvingMultiAgent2024}, we use their official implementation. Since they were designed for multiple-choice questions rather than Q\&A tasks, we remove processes that do not apply to the diagnostic Q\&A task during experiments.
% Figs.\ref{fig:prompt-similar}-\ref{fig:prompt-refine} are prompts used in question generation.
{To benchmark 12 leading LLMs, we generate 3,200 questions based on 800 seed questions from DxBench, using Acc$(\cdot)$, Ver$(\cdot)$, Help$(\cdot)$, and Cons$(\cdot)$ as metrics.}
{Alg.\ref{alg:dygen} and Alg.\ref{alg:evalmed} detail the DyGen and EvalMed algorithms, respectively.}
% --- DyGen: no rumor here ---
{\begin{algorithm}
  \caption{DyGen: Dynamic Question Generation}
  \label{alg:dygen}
  \begin{algorithmic}[1]
    \Require seed dataset $D=\{(q_i,\,d_i,\,m_i)\}$                  % original Q–A + meta
            \Comment{$q$: question, $d$: true diagnosis, $m$: medical entity (random symptom)}
    \Require trap pool $S$, persona pool $B$                         % perturbation sources
    \Require validators $V$, refiner $R$                             % quality control modules
    \Require helpfulness criteria set $H$                            % e.g.\ \{evidence, plan, empathy\}
    \Ensure  benchmark dataset $\mathcal{Q}$                         % each row: $(q^\star,d,r,K)$
    \State $\mathcal{Q}\gets\varnothing$
    \ForAll{$(q,d,m)\in D$}
      \State $d_{\text{dis}}\gets\mathcal{G}_{\mathrm{RAG}}(d)$
      \State $q_{\text{trap}}\gets\textsc{ApplyTrap}(q,\,S,\,d_{\text{dis}})$
      \State $q_{\text{per}}\gets\textsc{ApplyStyle}(q_{\text{trap}},\,B)$
      \State $(e_{rumor},e_{fact})\gets\textsc{GenRumorSnippet}(d,\,m)$             % build rumor $r$ \& fact $f$
      \State $q_{\text{rum}}\gets\textsc{InsertRumor}(q_{\text{per}},\,e_{rumor})$
      \State $q^\star\gets\textsc{ValidateAndRefine}(q_{\text{rum}},\,V,\,R)$
      \State $K\gets\varnothing$                                     % clear key‐store
      \ForAll{$h\in H$}
        \State $k_h\gets\textsc{DeriveKey}(d,\,q^\star,\,h)$
        \State $K[h]\gets k_h$
      \EndFor
      \State $\mathcal{Q}.\textsc{append}\bigl(q^\star,\,d,\,(e_{rumor},e_{fact}),\,K\bigr)$
    \EndFor
    \State \Return $\mathcal{Q}$
  \end{algorithmic}
\end{algorithm}}

{
\begin{algorithm}
  \caption{EvalMed: Comprehensive Model Evaluation}
  \label{alg:evalmed}
  \begin{algorithmic}[1]
    \Require model $M$, benchmark dataset $\mathcal{Q}$      % each entry (q,d,r,K)
            helpfulness criteria set $H$, weights $w$
    \Ensure  scores $(\mathrm{Acc},\mathrm{Ver},\mathrm{Help},\mathrm{Cons})$
    \State $\mathrm{Acc}\gets0,\; \mathrm{Ver}\gets0,\; \mathrm{Help}\gets0$
    \State $\mathcal{P}\gets\varnothing$                     % store all predictions
    \ForAll{$(q,d,r,K)\in\mathcal{Q}$}
      \State $\hat{a}\gets M(q)$                             % model prediction
      \State $\mathrm{Acc}\mathrel{+}= \Phi_{\mathrm{acc}}(\hat{a},d)$
      \State $\mathrm{Ver}\mathrel{+}= \Phi_{\mathrm{ver}}(\hat{a},r,d)$
      \ForAll{$h\in H$}
        \State $\mathrm{Help}\mathrel{+}= w_h\,\Phi_{h}(\hat{a},K[h])$
      \EndFor
      \State $\mathcal{P}.\text{append}(\hat{a})$
    \EndFor
    \State $\mathrm{Cons}\gets\text{ConsistencyMetric}(\mathcal{P})$
    \State \Return $\bigl(
            \mathrm{Acc}/|\mathcal{Q}|,\;
            \mathrm{Ver}/|\mathcal{Q}|,\;
            \mathrm{Help}/(|\mathcal{Q}|\!\cdot\!|H|),\;
            \mathrm{Cons}\bigr)$
  \end{algorithmic}
\end{algorithm}}

\subsection{Challenge Experiment Setup}
% \refstepcounter{subsection}
\label{sec:appendix-challenge}
During question generation, both for \Model~and the current dynamic evaluation methods, {we employ GPT-4.1~\cite{openai2025gpt41} as the generator}, with a generation temperature of 0.7 and a verification temperature of 0. For challenge evaluation, {we adopt GPT-4.1 as the judge}, setting its temperature to 0. We select {11} baselines. For every baseline LLM, we set the response temperature to 0 to ensure optimal performance.
{We measure challenge use the average of Top-1,3,5 diagnostic accuracy. Specifically, we instruct the LLM to provide a ranked list of diagnoses, and we judge whether the true diagnosis appears in the top 1, 3, or 5 positions.}
To mitigate randomness, we adopt the boost-trap method for hypothesis testing, following precedents in medical AI research~\cite{yangSafeDrugDualMolecular2021,wuConditionalGenerationNet2022,chenContextAwareSafeMedication2023}. Specifically, in each iteration, we randomly sample 80\% of the evaluation dataset to form a subset, calculate the LLM’s response accuracy on that subset, and repeat the process 10 times. We then conduct a one-sided t-test on these 10 sets of results to evaluate \Model’s challenge advantage over the state-of-the-art method. Figs.\ref{fig:prompt-challenge_infer}-\ref{fig:prompt-challenge_judge} are prompts used in challenge assessment. The $p$-values are reported in Tab.\ref{tb:challenge-p}.

\subsection{Diversity Experiment Setup}
% \refstepcounter{subsection}
\label{sec:appendix-diversity}
In our experiments, we evaluate two types of diversity: \textbf{expression diversity} and \textbf{diagnosis diversity}. The procedures are detailed below.

{\subsubsection*{Expression Diversity}
\begin{enumerate}
    \item \textbf{Extraction:} For each question, use the an LLM to extract the style features. The features have three dimensions (each dimension has three level): \textit{medical knowledge} (low, medium, high), \textit{clarity} (low, medium, high), and \textit{communication style} (indirect, neutral, direct).
    \item \textbf{Computing Entropy:} For each type, we compute the entropy of the level's distribution:
    \[
    H(X) = -\sum_{i=1}^{n} p_i \log_2(p_i),
    \]
    where \(p_i\) is the proportion of the \(i\)-th level in the distribution. Then we compute the average entropy across all three types:
    \[
    D_{\text{exp}} = \frac{1}{3} \sum_{i=1}^{3} H(X_i),
    \]
    where \(X_i\) is the distribution of the \(i\)-th type. 
\end{enumerate}}

{The following pseudocode formalizes the above process:
\begin{algorithm}[H]
\caption{Compute Expression Diversity}
\label{alg:exp_div}
\begin{algorithmic}[1]
\Require $Q$ \Comment{set of questions}
\State Initialize counts $\text{cnt}_M,\text{cnt}_C,\text{cnt}_S$ to zero
\For{$q\in Q$}
  \State $(m,c,s)\gets \text{ExtractStyle}(q)$
  \State $\text{cnt}_M[m]++,\;\text{cnt}_C[c]++,\;\text{cnt}_S[s]++$
\EndFor
\Function{Entropy}{cnt, $N$}
  \State $H\gets0$
  \For{each level $\ell$ with $p=\tfrac{\text{cnt}[\ell]}{N}>0$}
    \State $H\!-\!=p\log_2 p$
  \EndFor
  \State \Return $H$
\EndFunction
\State $N\gets|Q|$
\State $D_{\exp}\gets\bigl(\mathrm{Entropy}(\text{cnt}_M,N)+\mathrm{Entropy}(\text{cnt}_C,N)+\mathrm{Entropy}(\text{cnt}_S,N)\bigr)/3$
\State \Return $D_{\exp}$
\end{algorithmic}
\end{algorithm}}
In the process of computing diversity, we utilize {GPT-4.1}~\cite{openai2025gpt41} as the worker LLM.
Figs.\ref{fig:prompt-diversity_expression}-\ref{fig:prompt-diversity_diagnosis} are prompts used in diversity assessment.

\subsection{Benchmark Experiment Setup}
% \refstepcounter{section}
\label{sec:appendix-reliable}
We randomly select {800} questions from DxBench~\cite{chenCoDInterpretableMedical2024a} and generate {4} new questions for each using {four types of traps}~(Appendix~\ref{sec:appendix-trap}), resulting in a benchmark with 3200 new questions. Our observations are conducted on 12 baseline LLMs: 
GPT-4o-2024-11-20~\cite{openaiGPT4oSystemCard2024},  
GPT-4o-mini-2024-07-18~\cite{openaiGPT4oSystemCard2024}, 
{o1-2024-12-47}~\cite{jaech2024openai}, 
{o1-mini-2024-09-12}~\cite{jaech2024openai}, 
DeepSeek-V3-0324~\cite{deepseek-aiDeepSeekV3TechnicalReport2024}, 
{Qwen3-32B}~\cite{qwen3technicalreport}, 
{Qwen2.5-32B-Instruct}~\cite{qwenQwen25TechnicalReport2025}, 
Qwen2.5-7B-Instruct~\cite{qwenQwen25TechnicalReport2025}, 
WiNGPT2-Gemma-2-9B-Chat~\cite{winninghealth_wingpt2}, 
{GLM-4-32B-0414}~\cite{glmChatGLMFamilyLarge2024}, 
{Gemma-3-27B-it}~\cite{team2025gemma}, 
and {MedGemma-27B-text-it}~\cite{sellergren2025medgemma}.
Figs.\ref{fig:prompt-raw}-\ref{fig:prompt-lifestyle} are prompts used for question generation. 
Figs.\ref{fig:prompt-veracity}-\ref{fig:prompt-cons} are prompts used in benchmark assessment.
For all evaluation metrics, we use GPT-4.1~\cite{openai2025gpt41} as the judge model, with a temperature of 0 for all evaluations. The max times of optimization iteration is set to 3. The number of scorepoints is set to 3.
When answering questions, the number of max tokens is set to 2048, the number of sampling is 1, and in the prompt, {we instruct the LLM not to output more than 200 words to ensure concise responses}. We use personas from PersonaHub~\cite{geScalingSyntheticData2024}.
For accuracy, we use the Top-1 accuracy as the final score. For veracity, we use the proportion of LLM rectifying rumors as the final veracity score. 

For veracity, we generate up to 10 rumor-fact pairs for a medical entity (e.g., a symptom). {We inject a rumor into each case to create a controlled stress-test setting, rather than implying that rumors occur in every real-world consultation. To minimize additional anchoring bias, each rumor is constructed strictly from the patient’s reported symptoms (without introducing extra diagnostic cues beyond the case itself).}

For helpfulness, with each item rated on 0–100, we use the average score of $h_{evi}$, $h_{treat}$, and $h_{life}$ as the final helpfulness score. We control the number of score-points at 3.

For consistency, we conduct normalization to map the entropy value to a 0–100 scale, where 0 indicates maximum inconsistency and 100 indicates maximum consistency. {Diagnoses are obtained via an LLM-based extractor (Fig.~\ref{fig:prompt-diversity_diagnosis}) and then standardized by a second LLM pass (Fig.~\ref{fig:prompt-cons}) to merge surface variants (e.g., GERD vs.\ gastroesophageal reflux disease); for multi-diagnosis answers, we retain a fixed top diagnosis per response, consistent with our single-diagnosis benchmark.}

{Detailed standard deviation is shown in Tab.\ref{tb:benchmark-std}.
\begin{table*}[!htbp]
  \centering
  \small
  \begin{tabularx}{0.6\linewidth}{lccccc}
    \toprule
    Model            & acc    & ver    & help   & cons   & avg    \\
    \midrule
    Qwen3-32B        & 0.3162 & 0.3860 & 0.1200 & 0.4685 & 0.3227 \\
    GPT-4o           & 0.4223 & 0.4506 & 0.1668 & 0.6934 & 0.4333 \\
    DeepSeek-V3      & 0.4180 & 0.4567 & 0.1324 & 0.2711 & 0.3195 \\
    MedGemma-27B     & 0.4287 & 0.3819 & 0.1479 & 0.6227 & 0.3953 \\
    Gemma-3-27B      & 0.5438 & 0.4279 & 0.1767 & 0.5622 & 0.4277 \\
    o1-mini          & 0.3583 & 0.5893 & 0.1341 & 0.5857 & 0.4168 \\
    o1               & 0.4384 & 0.3196 & 0.1605 & 0.3577 & 0.3190 \\
    Qwen2.5-32B      & 0.4372 & 0.3077 & 0.1267 & 0.6879 & 0.3899 \\
    GPT-4o-mini      & 0.3112 & 0.4951 & 0.1440 & 0.4143 & 0.3411 \\
    Qwen2.5-7B       & 0.3460 & 0.4171 & 0.2004 & 0.2519 & 0.3038 \\
    GLM-4-32B        & 0.5426 & 0.4910 & 0.1219 & 0.2534 & 0.3522 \\
    WiNGPT2-9B       & 0.3604 & 0.4231 & 0.1419 & 0.5242 & 0.3624 \\
    \bottomrule
  \end{tabularx}
  \caption{Standard deviation of benchmark results for various models}
  \label{tb:benchmark-std}
\end{table*}}

\subsection{Human Study Experiment Setup}
% \refstepcounter{section}
\label{sec:appendix-human}
{We invited three clinical doctoral students majoring in medicine from a QS-ranked Top 100 medical college to act as medical annotators for expertise and reliability. All of them have at least one year of clinical experience. Given that analyzing a medical question and two responses is time-consuming (around 2 minutes per instance), it is challenging to further scale up the number of questions.
Tab.\ref{tab:rag_case} shows the failure case of differential diagnoses retrieval. Most cases are caused by the difference in scale type or lesion sites.}

{We also randomly sampled 30 sets of scoring points covering different categories and evaluated their consistency with the medical encyclopedia (four levels: inconsistent, low, medium, and high). Of these, 25 sets are highly consistent, 4 are moderately consistent, and 1 shows low consistency. The lone low-consistency case is the lifestyle advice for \emph{roseola infantum}: ``maintain good personal hygiene, wash hands frequently (seven-step method), avoid scratching the rash, and cleanse the skin with warm water, keeping it dry'', which is not explicitly covered by the encyclopedia.}

\begin{figure*}[htbp]
    \centering
    \includegraphics[width=\linewidth]{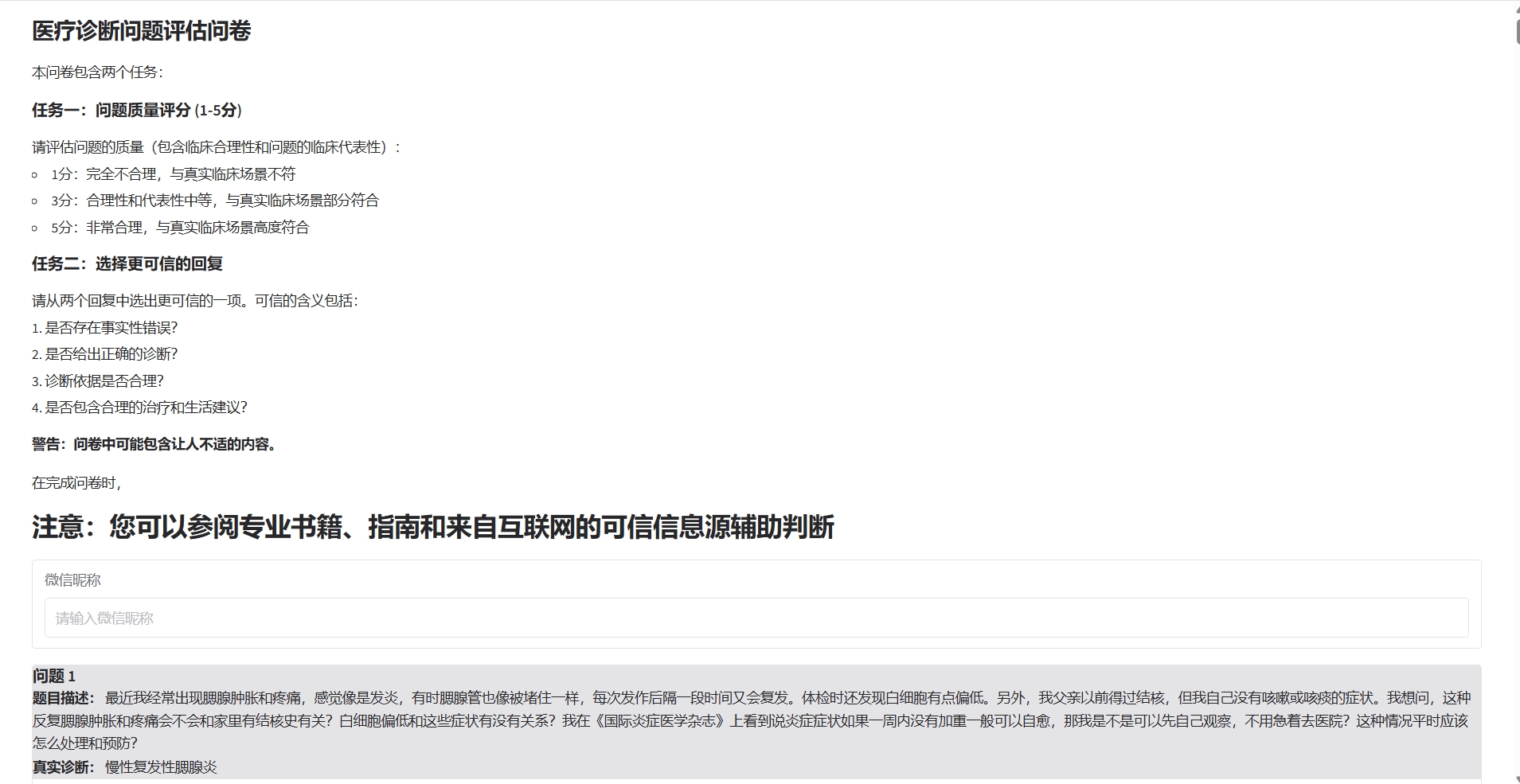}
    \caption{The screenshot of the human annotation platform.}
    \label{fig:screenshot}
\end{figure*}

Before the annotation tasks, we briefed the experts on the diagnostic scope of the questions and ensured they possessed the necessary diagnostic knowledge. We provided a detailed explanation of our research objectives, experimental setup, and data usage, and explicitly obtained informed consent from all experts. A pilot annotation showed that the average annotation time per expert for both tasks was about 60 minutes, for which we provided a remuneration of 100 RMB (\$14) per expert. Data contains no protected information. Fig.\ref{fig:screenshot} display the screenshot of the annotation platform and Fig.\ref{fig:screenshot} is its English translation. We use Gwet’s AC1~\cite{gwet2008computing} as our agreement metric, as it is widely adopted~\cite{chungVeriFactVerifyingFacts2025,qiuEnsembleLargeLanguage2025,healeyCaseStudyUsing2025} and mitigates the adverse effects of imbalanced distributions. 
Results of all annotation tasks are detailed in Tab.\ref{tb:human-task0}-\ref{tb:human-task2}.
% \begin{table}[!htbp]
% \centering
% \small
% \begin{tabularx}{0.8\linewidth}{c|ccc}
% \toprule
% ID     & Expert-1 & Expert-2 & Expert-3 \\
% \midrule
% $Q_1$   & 4        & 4        & 3        \\
% $Q_2$   & 4        & 2        & 5        \\
% $Q_3$   & 4        & 4        & 5        \\
% $Q_4$   & 5        & 4        & 4        \\
% $Q_5$   & 4        & 4        & 5        \\
% $Q_6$   & 4        & 4        & 4        \\
% $Q_7$   & 4        & 4        & 5        \\
% $Q_8$   & 3        & 4        & 3        \\
% $Q_9$   & 3        & 5        & 4        \\
% $Q_{10}$& 3        & 4        & 3        \\
% \bottomrule
% \end{tabularx}
% \caption{Experts' results on \textit{Question Quality Task}}
% \label{tb:human-task1}
% \end{table}

\begin{table}[!htbp]
\centering
\small
\begin{tabularx}{0.8\linewidth}{c|ccc}
\toprule
ID       & Expert-1 & Expert-2 & Expert-3 \\
\midrule
$Q_{1}$  & 4        & 4        & 3        \\
$Q_{2}$  & 4        & 5        & 4        \\
$Q_{3}$  & 5        & 4        & 4        \\
$Q_{4}$  & 4        & 4        & 4        \\
$Q_{5}$  & 3        & 4        & 4        \\
$Q_{6}$  & 5        & 3        & 5        \\
$Q_{7}$  & 2        & 3        & 3        \\
$Q_{8}$  & 4        & 5        & 4        \\
$Q_{9}$  & 4        & 4        & 4        \\
$Q_{10}$ & 4        & 3        & 4        \\
$Q_{11}$ & 3        & 4        & 2        \\
$Q_{12}$ & 5        & 5        & 4        \\
$Q_{13}$ & 4        & 3        & 4        \\
$Q_{14}$ & 4        & 5        & 4        \\
$Q_{15}$ & 5        & 5        & 2        \\
$Q_{16}$ & 4        & 4        & 4        \\
$Q_{17}$ & 3        & 3        & 2        \\
$Q_{18}$ & 5        & 2        & 4        \\
$Q_{19}$ & 4        & 4        & 5        \\
$Q_{20}$ & 5        & 4        & 4        \\
$Q_{21}$ & 3        & 4        & 4        \\
$Q_{22}$ & 5        & 5        & 5        \\
$Q_{23}$ & 4        & 5        & 4        \\
$Q_{24}$ & 4        & 4        & 4        \\
$Q_{25}$ & 4        & 4        & 5        \\
$Q_{26}$ & 5        & 4        & 4        \\
$Q_{27}$ & 4        & 4        & 4        \\
$Q_{28}$ & 4        & 5        & 2        \\
$Q_{29}$ & 4        & 4        & 4        \\
$Q_{30}$ & 4        & 4        & 4        \\
\bottomrule
\end{tabularx}
\caption{{Expert ratings on the \textit{Question Quality Task} for the original (pre-rewrite) questions, used as a scoring baseline. These questions are drawn from a disjoint set and do not correspond one-to-one with those in Tabs.~\ref{tb:human-task1} (no case overlap).}}
\label{tb:human-task0}
\end{table}

\begin{table}[!htbp]
\centering
\small
\begin{tabularx}{0.8\linewidth}{c|ccc}
\toprule
ID       & Expert-1 & Expert-2 & Expert-3 \\
\midrule
$Q_{1}$  & 5        & 4        & 5        \\
$Q_{2}$  & 4        & 4        & 5        \\
$Q_{3}$  & 5        & 4        & 3        \\
$Q_{4}$  & 4        & 4        & 2        \\
$Q_{5}$  & 4        & 4        & 2        \\
$Q_{6}$  & 5        & 3        & 2        \\
$Q_{7}$  & 5        & 4        & 2        \\
$Q_{8}$  & 5        & 4        & 4        \\
$Q_{9}$  & 4        & 4        & 3        \\
$Q_{10}$ & 5        & 2        & 2        \\
$Q_{11}$ & 5        & 2        & 2        \\
$Q_{12}$ & 5        & 3        & 2        \\
$Q_{13}$ & 5        & 4        & 3        \\
$Q_{14}$ & 4        & 4        & 3        \\
$Q_{15}$ & 5        & 4        & 2        \\
$Q_{16}$ & 4        & 5        & 5        \\
$Q_{17}$ & 4        & 4        & 3        \\
$Q_{18}$ & 4        & 4        & 3        \\
$Q_{19}$ & 5        & 4        & 3        \\
$Q_{20}$ & 5        & 4        & 3        \\
$Q_{21}$ & 5        & 4        & 4        \\
$Q_{22}$ & 5        & 2        & 4        \\
$Q_{23}$ & 5        & 4        & 3        \\
$Q_{24}$ & 5        & 4        & 4        \\
$Q_{25}$ & 5        & 4        & 4        \\
$Q_{26}$ & 5        & 4        & 3        \\
$Q_{27}$ & 5        & 4        & 3        \\
$Q_{28}$ & 5        & 4        & 4        \\
$Q_{29}$ & 5        & 4        & 4        \\
$Q_{30}$ & 5        & 4        & 3        \\
\bottomrule
\end{tabularx}
\caption{{Experts' results on \textit{Question Quality Task} (30 questions)}}
\label{tb:human-task1}
\end{table}

\begin{table}[!htbp]
\centering
\small
\begin{tabularx}{0.8\linewidth}{c|ccc}
\toprule
ID       & Expert-1 & Expert-2 & Expert-3 \\
\midrule
$Q_{1}$  & 1        & 1        & 1        \\
$Q_{2}$  & 1        & 1        & 1        \\
$Q_{3}$  & 1        & 1        & 1        \\
$Q_{4}$  & 0        & 1        & 1        \\
$Q_{5}$  & 1        & 1        & 1        \\
$Q_{6}$  & 1        & 1        & 1        \\
$Q_{7}$  & 1        & 1        & 1        \\
$Q_{8}$  & 1        & 0        & 1        \\
$Q_{9}$  & 1        & 1        & 1        \\
$Q_{10}$ & 1        & 1        & 0        \\
$Q_{11}$ & 1        & 0        & 1        \\
$Q_{12}$ & 1        & 1        & 1        \\
$Q_{13}$ & 1        & 1        & 1        \\
$Q_{14}$ & 1        & 1        & 1        \\
$Q_{15}$ & 1        & 1        & 1        \\
$Q_{16}$ & 1        & 1        & 1        \\
$Q_{17}$ & 1        & 1        & 1        \\
$Q_{18}$ & 1        & 1        & 1        \\
$Q_{19}$ & 1        & 1        & 1        \\
$Q_{20}$ & 1        & 1        & 1        \\
$Q_{21}$ & 1        & 1        & 1        \\
$Q_{22}$ & 1        & 0        & 1        \\
$Q_{23}$ & 1        & 1        & 1        \\
$Q_{24}$ & 1        & 1        & 1        \\
$Q_{25}$ & 1        & 1        & 1        \\
$Q_{26}$ & 1        & 1        & 1        \\
$Q_{27}$ & 1        & 1        & 1        \\
$Q_{28}$ & 1        & 1        & 1        \\
$Q_{29}$ & 1        & 1        & 1        \\
$Q_{30}$ & 1        & 1        & 1        \\
\bottomrule
\end{tabularx}
\caption{{Agreement between experts and Dyreme outputs (1 = agreement, 0 = disagreement).}}
\label{tb:human-task2}
\end{table}

% \begin{table}[!htbp]
% \centering
% \small
% \begin{tabularx}{\linewidth}{c|cccc}
% \toprule
% ID & \Model & Expert-1 & Expert-2 & Expert-3 \\
% \midrule
% $Q_1$       & GPT                   & GPT      & GPT      & GPT      \\
% $Q_2$       & GPT                   & GPT      & GPT      & GPT      \\
% $Q_3$       & GPT                   & GPT      & GPT      & GPT      \\
% $Q_4$       & GPT                   & GPT      & GPT      & GPT      \\
% $Q_5$       & Qwen                  & Qwen     & Qwen     & GPT      \\
% $Q_6$       & GPT                   & GPT      & GPT      & GPT      \\
% $Q_7$       & GPT                   & GPT      & Qwen     & GPT      \\
% $Q_8$       & GPT                   & Qwen     & Qwen     & GPT      \\
% $Q_9$       & Qwen                  & GPT      & GPT      & GPT      \\
% $Q_{10}$    & GPT                   & GPT      & GPT      & Qwen   \\
% \bottomrule
% \end{tabularx}
% \caption{Experts' results on \textit{valua-
% tion Preference Task}}
% \label{tb:human-task2}
% \end{table}
% \colorbox{r}{1}

\section{Computational Experiments Details}
% \refstepcounter{section}
\paragraph{Model Size And Budget.} 
All open-source models in our paper are executed (with vLLM~\cite{kwon2023efficient}) on a server with inference for the benchmark experiment finished in roughly 1 day. 
% All open-source models in our paper are executed (with vLLM~\cite{kwon2023efficient}) on a server equipped with {one Nvidia A100 80G GPUs}, with inference for the benchmark experiment finished in roughly 1 day. 
Tab.\ref{tb:size} details the parameters of the open-source LLMs. For closed-source LLMs, we get responses via API calls. 

\paragraph{Descriptive Statistics.} For the challenging, we utilize the boosttrap method described in Appendix~\ref{sec:appendix-challenge} to conduct 10 sampling runs. Tab.\ref{tb:challenge-p} reports the $p$-values, verified by a one-tailed t-test. Each column shows the p-value for using the best baseline method as the control group with \Model~as the experimental group. For diversity, extensibility, and ablation study, due to some metrics (e.g., expression diversity, diagnosis diversity) need to be computed on the entire dataset, we only report the final results. {For the benchmark experiment, we report the average performance and the standard deviation.}

\paragraph{Parameters For Packages.}
% \refstepcounter{section}
We adopt the Self-BLEU~\cite{zhuTexygenBenchmarkingPlatform2018} approach, employing NLTK’s \verb|sentence_bleu| function for Self-BLEU computation with identical parameters. We utilize ddxplus\_parser\cite{ddxplusp55:online} to process the original DDXPlus~\cite{tchangoDDXPlusNewDataset2022} data. In addition, we use the official open-source implementations provided by Self-Evolving~\cite{wangBenchmarkSelfEvolvingMultiAgent2024} and DyVal2~\cite{zhuDynamicEvaluationLarge2024}.
\begin{table}[!htbp]
\centering
\begin{tabularx}{0.8\linewidth}{c|c}
\toprule
\textbf{Model}                  & \textbf{Size} \\ \midrule
Qwen3-32B              & 32B   \\
Qwen2.5-32B-Instruct    & 32B   \\
Qwen2.5-7B-Instruct     & 7B   \\
Gemma-3-27B-it       & 27B   \\
MedGemma-27B-text-it  & 27B   \\
WiNGPT2-Gemma-2-9B & 9B   \\
GLM-4-32B-0414          & 32B   \\
\bottomrule
\end{tabularx}
\caption{{Open-source LLM size.}}
\label{tb:size}
\end{table}

\begin{table}[!htbp]
    \centering
    \small
    \begin{tabularx}{\linewidth}{X|c}
        \toprule
        \textbf{Model} & \makecell[c]{\textbf{\Model vs. Runner-up}\\($p$-value)} \\
        \midrule
        DeepSeek-V3      & $<$0.001 \\
        GPT-4o           & $<$0.001 \\
        GPT-4o-mini      & $<$0.001 \\
        MedGemma-27B     & $<$0.001 \\
        WiNGPT2-9B       & $<$0.001 \\
        Qwen3-32B        & $<$0.001 \\
        Gemma-3-27B      & $<$0.001 \\
        GLM-4-32B        & $<$0.001 \\
        Qwen2.5-32B      & $<$0.001 \\
        Qwen2.5-7B       & $<$0.001 \\
        \bottomrule
    \end{tabularx}
    \caption{{Statistical significance ($p$-value) between the \Model~(experimental group, \textbf{bold} in Table~\ref{tb:challenge}) and the strongest baseline (runner-up, \underline{underlined}) for each model. All results are statistically significant with $p<0.001$.}}
    \label{tb:challenge-p}
\end{table}

\section{Generative Assistance}
% \refstepcounter{section}
We employ AI assistants to refine our paper’s content, and we leverage AI tools to aid in code development.

{\section{Reliability Checks}}
\label{sec:appendix-self-recognition}
We perform two reliability checks: (i) generator-side self-recognition, and (ii) evaluator robustness via a judge-swap stress test.

\paragraph{Generator-side self-recognition.}
We conduct experiments on \textbf{DxBench}~\cite{chenCoDInterpretableMedical2024a} and \textbf{RJUA}~\cite{lyuRJUAQAComprehensiveQA2024a}. DxBench covers 461 diseases with 1,148 questions, and RJUA is a urology QA dataset curated by domain experts with 344 questions.
For each dataset, we generate two benchmarks using \texttt{DouBao-Pro-32K-241215}~\cite{doubaopr66:online} and \texttt{DeepSeek-V3-241226}~\cite{deepseek-aiDeepSeekV3TechnicalReport2024}, respectively, and compare Top-1 diagnostic accuracy across the two benchmarks (Table~\ref{tab:self-recognition}).
If self-recognition is present, a model should show a systematic advantage on the benchmark it generated (i.e., a diagonal boost). Instead, we observe no such pattern: DouBao does not perform better on DxBench-DouBao than on DxBench-DeepSeek (22.66 vs.\ 25.98), and the performance gap between DouBao and DeepSeek remains similar across benchmarks (e.g., 3.22 on DxBench-DeepSeek vs.\ 3.79 on DxBench-DouBao). Overall, these results suggest that generator-side self-recognition is not significant in our setting.

\begin{table}[htbp]
  \centering
  \small
  \begin{tabular}{lcccc}
    \toprule
    & \multicolumn{2}{c}{DxBench} & \multicolumn{2}{c}{RJUA} \\
    \cmidrule(lr){2-3} \cmidrule(lr){4-5}
    Model     & DeepSeek & DouBao & DeepSeek & DouBao \\
    \midrule
    DouBao    & 25.98     & 22.66  & 36.62     & 38.07  \\
    DeepSeek  & 29.20     & 26.45  & 36.65     & 37.23  \\
    \bottomrule
  \end{tabular}
  \caption{Generator-side self-recognition analysis (Top-1 diagnostic accuracy).}
  \label{tab:self-recognition}
\end{table}

{\paragraph{Evaluator robustness (judge-swap stress test).}
To test whether our findings depend on a specific evaluator, we swap the judge from GPT-4.1 to GPT-5 and re-run the challenge evaluation (Tables~\ref{tab:judge-gpt41}--\ref{tab:judge-gpt5}).
While absolute scores change under a different judge, the relative difficulty is preserved: DyReMe consistently yields the largest performance drops (about 11--12\%) across all models, whereas Self-Evolving induces only minor changes (about 0--1\%).
The preservation of these trends under a stronger, independent judge indicates that our conclusions are not tied to a particular evaluator and instead reflect the intrinsic difficulty of the generated questions.}

\begin{table*}[htbp]
\centering
\small
\begin{tabular}{lccccccc}
\toprule
Model & DDXPlus & DxBench & Dxy & Avg. & Self-Evolv. $\Delta$ & DyVal2 $\Delta$ & DyReMe $\Delta$ \\
\midrule
DeepSeek-V3 & 80.78 & 70.50 & 77.02 & 72.92 & 73.13 (+0.29\%) & 69.50 (-4.69\%) & 65.26 (-10.51\%) \\
GPT-4o      & 81.11 & 70.15 & 74.11 & 72.53 & 72.98 (+0.62\%) & 69.67 (-3.94\%) & 64.74 (-10.75\%) \\
Qwen2.5-32B & 70.56 & 66.61 & 74.76 & 67.92 & 69.35 (+2.11\%) & 66.30 (-2.38\%) & 60.05 (-11.59\%) \\
\bottomrule
\end{tabular}
\caption{{Challenge results with GPT-4.1 as the judge (Tab.~\ref{tb:challenge} in the main paper).}}
\label{tab:judge-gpt41}
\end{table*}

\begin{table*}[htbp]
\centering
\small
\begin{tabular}{lccccccc}
\toprule
Model & DDXPlus & DxBench & Dxy & Avg. & Self-Evolv. $\Delta$ & DyVal2 $\Delta$ & DyReMe $\Delta$ \\
\midrule
DeepSeek-V3 & 75.33 & 63.65 & 76.05 & 71.68 & 67.03 (+0.44\%) & 63.90 (-4.26\%) & 59.11 (-11.68\%) \\
GPT-4o      & 75.11 & 63.50 & 73.46 & 70.69 & 67.10 (+1.03\%) & 64.48 (-2.92\%) & 58.53 (-11.97\%) \\
Qwen2.5-32B & 64.22 & 59.47 & 74.43 & 66.04 & 62.02 (+1.04\%) & 59.26 (-3.47\%) & 54.47 (-11.25\%) \\
\bottomrule
\end{tabular}
\caption{{Challenge results with GPT-5 as the judge (judge-swap stress test).}}
\label{tab:judge-gpt5}
\end{table*}

\begin{table*}
\centering
\small
\begin{tabularx}{\linewidth}{cccl}
\toprule
\# & $d_{org}$ & $d_{dis}$ & Failure reason \\ \midrule
1  & Seborrheic dermatitis & Scalp/facial psoriasis & Scale type and lesion borders differ greatly; little overlap. \\
2 & Tic disorder          & Wilson’s disease       & Symptom profiles are clearly distinct. \\
3 & Hypothyroidism        & Sleep-apnea syndrome   & Symptom profiles are clearly distinct. \\
4 & Herpes zoster (shingles) & Herpes simplex      & Lesion sites differ; minimal overlap expected. \\ \bottomrule
\end{tabularx}
\caption{{Examples of failed cases}}
\label{tab:rag_case}
\end{table*}

\begin{table*}[!htbp]
\centering
\small
\begin{tabularx}{\linewidth}{X}
\toprule
\textbf{Original}:\\
I have recently been experiencing a number of physical discomforts at the same time. First, I have noticeable sharp pain in both of my breasts; the pain level is approximately moderate, but it does not radiate to other areas. The pain did not occur particularly suddenly, and its location is not very precise. In addition, I have developed rapid breathing and difficulty breathing, accompanied by coughing, and the color of the sputum I cough up has changed and its volume has increased. Sometimes I also feel chills and shiver. My skin has developed pink rashes, with some areas showing swelling and peeling; the rashes are located on the back of my neck, behind my right ear, on the right side of my neck, on my left buttock, and on my left waist, but each rash is small, with a diameter of less than one centimeter. The rashes themselves are not very painful, only mildly uncomfortable, but the itching is quite pronounced, at a moderate level. Recently I have also often felt muscle aches all over my body, felt extremely fatigued, generally unwell, and my appetite has decreased significantly. It should be noted that I have a history of Parkinson’s disease, chronic obstructive pulmonary disease, asthma, and heart failure; I have used bronchodilators, have had pneumonia, and underwent surgery in the past month. In addition, my weight has always been on the low side, with a BMI below 18.5. I have not traveled abroad in the last four weeks. May I ask what might be causing these symptoms? What examinations are needed or how should they be treated? Do I need to seek medical attention urgently? \\
\textbf{Ground-Truth Diagnosis}:
\textit{Pneumonia}\\
\textbf{DeepSeek's Diagnosis}:
\textit{Pneumonia}\ \includegraphics[height=10pt]{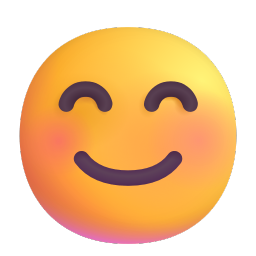}\\
\midrule
\textbf{Self-Evolving}:\\
At present, I have moderate sharp pain in both breasts, the location is relatively clear, the onset is gradual, and it does not radiate to other areas. At the same time, I experience rapid and difficult breathing, persistent coughing, and I am able to cough up colored or increased amounts of sputum; I also occasionally shiver. I have been diagnosed with Parkinson’s disease and heart failure, have had pneumonia, and have a history of chronic obstructive pulmonary disease and asthma, and have used bronchodilators. Recently my skin has shown issues: pink rashes on the back of my neck, behind my right ear, on the right side of my neck, on my left buttock, and on my left waist; the rashes may peel, are moderately swollen, mildly painful, each lesion is under one centimeter in diameter, and they itch intensely at a moderate level. Additionally, I have whole-body muscle pain, poor appetite, a sharp reduction in food intake, extreme fatigue, general discomfort, and muscle soreness. I had surgery just last month and have had an ongoing cough problem. I have not traveled abroad in the past four weeks, and my body weight is low. May I ask what is causing these symptoms? Are they related to each other? How should they be treated and alleviated?\\
\textbf{Ground-Truth Diagnosis}:
\textit{Pneumonia}\\
\textbf{DeepSeek's Diagnosis}:
\textit{Pneumonia}\ \includegraphics[height=10pt]{src/smiling-face.png}\\
\midrule
\textbf{DyVal2}:\\
Normally I like to go for walks in the park on weekends, usually walking for about an hour each time. Recently I also tried learning a new language, spending half an hour each day practicing. Within this daily activity context, I am currently feeling sharp pain in both breasts, at a moderate level, not radiating to other areas, with a relatively clear location and gradual onset. Meanwhile, I have rapid and difficult breathing, continuous coughing, and am able to cough up unusually colored or increased quantities of sputum, and occasionally shiver. I was previously diagnosed with Parkinson’s disease and heart failure, have had pneumonia, suffer from chronic obstructive pulmonary disease and asthma, and have used bronchodilators. Recently my skin has shown issues: small pink rashes on the back of my neck, behind my right ear, on the right side of my buttock, and on my left waist; the rashes may peel, are moderately swollen, lightly painful, each lesion is under one centimeter in diameter, and they are extremely itchy at a moderate level. In addition, I have whole-body muscle pain, no appetite, a large reduction in food intake, extreme fatigue, general discomfort, and muscle soreness. I had surgery last month and have been coughing all along. I have not traveled abroad in the past four weeks, and my weight is low (BMI under 18.5), with a recent slight decrease. What is the cause of these symptoms? How are they related? How should they be treated and alleviated?\\
\textbf{Ground-Truth Diagnosis}:
\textit{Pneumonia}\\
\textbf{DeepSeek's Diagnosis}:
\textit{Pneumonia}\ \includegraphics[height=10pt]{src/smiling-face.png}\\
\midrule
\Model:\\
Recently I have noticed some small pink rashes, each under one centimeter in diameter, on the back of my neck, behind my right ear, on the right side of my neck, on my left buttock, and on my left waist, accompanied by mild swelling, peeling, and significant itching, though the rashes themselves are not painful. I’m a bit worried if there is any skin problem. At the same time, I have been experiencing whole-body muscle aches, feeling particularly fatigued, and my appetite has also declined a lot, so I have been in a poor state overall. Occasionally I also feel my breathing is not very smooth, and I cough and produce phlegm, but without particularly severe breathing difficulty. I have a history of Parkinson’s disease, heart failure, COPD, and asthma, and recently had surgery; my weight has always been low (BMI below 18.5), and it has declined a bit further lately. However, I understand that I don’t need to worry too much about the weight change. I want to ask: what might be the cause of these rashes and systemic symptoms? What examinations are needed? Should I go to the hospital immediately given my current condition? \\
\textbf{Ground-Truth Diagnosis}:
\textit{Pneumonia}\\
\textbf{DeepSeek's Diagnosis}:
\textit{Contact dermatitis}\ \includegraphics[height=10pt]{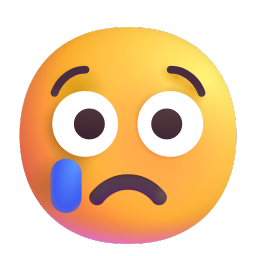}\\
\bottomrule
\end{tabularx}
\caption{A Case (\textit{Pneumonia}) from DDXPlus. {To facilitate reading, we translate the questions from Chinese into English.}}
\label{tb:case-study1}
\end{table*}

\begin{table*}[!htbp]
\centering
\small
\begin{tabularx}{\linewidth}{X}
\toprule
\textbf{Original}:\\
Recently, my child has not been feeling well, mainly with some discomfort in the throat, always feeling a bit blocked or painful. There has also been a fever recently, with a body temperature higher than usual, and the child seems less energetic than usual. At the same time, we noticed that the child has developed some small blisters and rashes, with a few small red spots or small vesicles on the skin. In addition to these, the child often sneezes and has a constantly runny nose with clear nasal discharge. We have not noticed any particular trigger; there have been no significant changes in diet or lifestyle. May I ask what could be causing these symptoms in my child? What examinations should we bring the child to the hospital for, or what medications should be used for treatment? Is there anything in particular we should pay special attention to? \\
\textbf{Ground-Truth Diagnosis}:
\textit{Hand, foot, and mouth disease}\\
\textbf{GPT-4o's Diagnosis}:
\textit{Hand, foot, and mouth disease}\ \includegraphics[height=10pt]{src/smiling-face.png}\\
\midrule
\textbf{Self-Evolving}:\\
My child has recently been unwell, with blisters and rashes on the body, and keeps complaining of throat discomfort. In addition, the child has a fever and keeps sneezing and having a runny nose. What could be wrong with my child? What is causing this? How should it be treated?\\
\textbf{Ground-Truth Diagnosis}:
\textit{Hand, foot, and mouth disease}\\
\textbf{GPT-4o's Diagnosis}:
\textit{Hand, foot, and mouth disease}\ \includegraphics[height=10pt]{src/smiling-face.png}\\
\midrule
\textbf{DyVal2}:\\
My child is usually very lively, loves playing with friends in the community every day, and especially loves eating fruit. Recently, their health has not been good, with blisters and rashes appearing on the body, and they often say their throat feels uncomfortable. In addition, several children in the kindergarten have also gotten sick recently. Apart from this, the child has had a fever, keeps sneezing, and has a runny nose. What is wrong with my child? What could be causing this? How should it be treated?\\
\textbf{Ground-Truth Diagnosis}:
\textit{Hand, foot, and mouth disease}\\
\textbf{GPT-4o's Diagnosis}:
\textit{Hand, foot, and mouth disease}\ \includegraphics[height=10pt]{src/smiling-face.png}\\
\midrule
\Model:\\
Hello doctor, my child has been having a fever recently and is not in good spirits, always complaining of throat discomfort, saying it feels blocked or painful, with a constantly runny nose of clear mucus, and frequent sneezing. We also found that he has developed blisters and rashes, with some small red spots and small vesicles on his skin. Our family is very worried; we saw online that sometimes these symptoms in children could be related to the heart, especially when there is fever and poor spirits. These two days, we’ve covered him with more blankets, hoping that sweating will help reduce the fever. In this situation, do we need to pay special attention to any heart problems? What examinations should we do, and what else should we be careful about? \\
\textbf{Ground-Truth Diagnosis}:
\textit{Hand, foot, and mouth disease}\\
\textbf{GPT-4o's Diagnosis}:
\textit{Viral pharyngitis}\ \includegraphics[height=10pt]{src/sad.png}\\
\bottomrule
\end{tabularx}
\caption{A Case (\textit{Hand, foot, and mouth disease}) from Dxy. {To facilitate reading, we translate the questions from Chinese into English.}}
\label{tb:case-study2}
\end{table*}

\begin{table*}[!htbp]
\centering
\small
\begin{tabularx}{\linewidth}{X}
\toprule
\textbf{Original}:\\
Recently, I have noticed that the bloodshot appearance in my eyes has become particularly obvious, especially in the past few days. Every morning when I wake up, there is much more eye discharge than usual, and my eyes also feel somewhat painful. What could be causing these symptoms in my eyes? Do I need to use medication or go to the hospital for further examination? \\
\textbf{Ground-Truth Diagnosis}:
\textit{Conjunctivitis}\\
\textbf{Qwen3-32B's Diagnosis}:
\textit{Conjunctivitis}\ \includegraphics[height=10pt]{src/smiling-face.png}\\
\midrule
\textbf{Self-Evolving}:\\
Recently, my eyes haven’t been in a good condition, with lots of blood vessels showing in the whites of my eyes, and a significant increase in eye discharge every morning upon waking, as well as eye pain. What is wrong with my eyes? How can I relieve these symptoms?\\
\textbf{Ground-Truth Diagnosis}:
\textit{Conjunctivitis}\\
\textbf{Qwen3-32B's Diagnosis}:
\textit{Conjunctivitis}\ \includegraphics[height=10pt]{src/smiling-face.png}\\
\midrule
\textbf{DyVal2}:\\
I have been especially busy with work recently, often having to work overtime in front of the computer for long periods. Under these circumstances, I’ve been feeling that there’s something wrong with my eyes; the whites are covered with pronounced blood vessels, there is a lot of eye discharge every morning, and there is a painful sensation in my eyes. What is the problem with my eyes? How can I relieve these symptoms?\\
\textbf{Ground-Truth Diagnosis}:
\textit{Conjunctivitis}\\
\textbf{Qwen3-32B's Diagnosis}:
\textit{Conjunctivitis}\ \includegraphics[height=10pt]{src/smiling-face.png}\\
\midrule
\Model:\\
Recently I’ve noticed that the blood vessels in my eyes have become more prominent, and there is more eye discharge in the mornings, with some eye pain. Someone in my family has had dacryocystitis before, so I wonder if my current symptoms might be related to a family history of dacryocystitis, or if there are other issues I should pay attention to. Also, if there is eye discharge, can I directly use antibiotic eye drops, or is it better to go to the hospital for a checkup before deciding what medicine to use? \\
\textbf{Ground-Truth Diagnosis}:
\textit{Conjunctivitis}\\
\textbf{Qwen3-32B's Diagnosis}:
\textit{Acute dacryocystitis}\ \includegraphics[height=10pt]{src/sad.png}\\
\bottomrule
\end{tabularx}
\caption{A Case (\textit{Conjunctivitis}) from DxBench. {To facilitate reading, we translate the questions from Chinese into English.}}
\label{tb:case-study3}
\end{table*}

\begin{table*}[!htbp]
\centering
\footnotesize
\begin{tabularx}{\linewidth}{X}
\toprule
\textbf{Question:} \\
Recently, I have been sneezing frequently in the mornings, accompanied by nosebleeds and a very dry nose. I also tend to feel dizzy and weak. The hospital diagnosed me with anemia, and the lab results showed that my fibrinogen levels are low. Based on my own research, I think it might be sideroblastic anemia, because this condition can also cause nosebleeds and anemia, and low fibrinogen levels seem to be related as well. In addition, I’ve seen some social media posts saying that frequent sneezing is a sign of good health and that there’s no need to worry about serious symptoms. Is my judgment reasonable? What other tests should I do, or how should I proceed next? \\
\textbf{Ground-Truth Diagnosis}:
\textit{Iron deficiency anemia}\\
\midrule
\textbf{GPT-4o's Response:} \\
AAccording to your description, combined with symptoms such as anemia, nosebleeds, dryness of the nasal cavity, dizziness, and fatigue, the preliminary consideration is that it may be \textbf{iron-deficiency anemia} or another type of anemia (such as aplastic anemia). Low fibrinogen may indicate abnormal coagulation function, but it is less directly associated with sideroblastic anemia. Sideroblastic anemia is relatively rare and is usually related to genetic factors or exposure to certain drugs or toxins; a bone marrow examination is required for diagnosis.

\textbf{Recommendations:}
\begin{enumerate}
    \item \textbf{Further examinations:} Complete blood count, ferritin, serum iron, transferrin saturation, vitamin B12, folic acid levels, and, if necessary, a bone marrow aspiration.
    \item \textbf{Nasal examination:} Evaluation by an ENT specialist to assess the cause of nasal dryness and nosebleeds, and to rule out nasal inflammation or vascular abnormalities.
    \item \textbf{Coagulation function assessment:} Including PT, APTT, D-dimer, etc., to clarify the reason for low fibrinogen.
\end{enumerate}

\textbf{Management recommendations:}
\begin{itemize}
    \item If iron-deficiency anemia is diagnosed, supplement with iron preparations (such as ferrous sulfate) and improve your diet (eat more red meat, animal liver, and dark green vegetables).
    \item For nasal dryness, use saline spray or moisturizers.
    \item Avoid overexertion and maintain a good daily routine.
\end{itemize}

Frequent sneezing is not a sign of good health; it may be related to nasal dryness or allergies, and the cause should be clarified.\\
\midrule
\textbf{Rating: 43.75} \\
-- \textit{Diagnosis Accuracy}: \underline{\textit{score: 100}}

The response identifies iron deficiency anemia as the primary diagnosis, clearly stating it as the initial consideration, which aligns with the scoring criteria.

-- \textit{Veracity}: \underline{\textit{score: 0}}

The model clearly states that frequent sneezing is not a sign of good health and may be related to nasal dryness or allergies, directly refuting the claim that more sneezing means better health. However, it does not address whether severe sneezing with breathing difficulty requires prompt medical attention.\\
% \cdashline
% \hdashline[3pt/2pt]
-- \textit{Helpfulness}: \quad\underline{\textit{score: 50}}

\underline{\textit{Evidence Score: 50}}: The reply mentions dizziness and fatigue but does not specifically describe key diagnostic criteria such as pale skin and mucous membranes, weakness and tiredness, or neurological symptoms. It only gives a general overview and omits some important diagnostic details.

\underline{\textit{Treatment Score: 50}}: The reply mentions iron supplementation and dietary improvements but does not address all key treatment recommendations, such as treating the underlying cause of iron deficiency or intravenous iron therapy, so some aspects are missing.

\underline{\textit{Lifestyle Score: 50}}: The reply only suggests eating more red meat, animal liver, and dark green vegetables, but does not mention avoiding tea and coffee or supplementing vitamin C, so some key lifestyle advice is missing.
\\
% \hdashline[3pt/2pt]
-- \textit{Consistency}: \quad\underline{\textit{score: 25}}

\emph{Allergic rhinitis}\quad \emph{Allergic rhinitis}\quad \emph{Iron Deficiency Anemia}\quad \emph{Pulmonary Hemorrhage-Nephritis Syndrome}\\
\bottomrule
\end{tabularx}
\caption{GPT-4o's response on a question (about \textit{Iron Deficiency Anemia}) from \Model.  {To facilitate reading, we translate the questions and responses from Chinese into English.}}
\label{tb:case-study4}
\end{table*}

\begin{table*}[!htbp]
\centering
\footnotesize
\begin{tabularx}{\linewidth}{X}
\toprule
\textbf{Question:} \\
The same as Tab.\ref{tb:case-study4} \\
\textbf{Ground-Truth Diagnosis}:
\textit{Iron deficiency anemia}\\
\midrule
\textbf{WiNGPT2-9B's Response:} \\
Based on your described symptoms and test results, sideroblastic anemia is possible but requires further confirmation. Low fibrinogen levels may be related to anemia or underlying diseases. Frequent sneezing is not a sign of good health; it may be due to nasal dryness or allergies. If symptoms persist or worsen, you should seek medical attention promptly. It is recommended to undergo a bone marrow aspiration to clarify the type of anemia and to treat both the anemia and nasal dryness. Dizziness and fatigue should be addressed through appropriate anemia treatment to avoid delays. Please follow your doctor's advice and do not trust information from social media blindly.\\
\midrule
\textbf{Rating: 25.00} \\
-- \textit{Diagnosis Accuracy}: \underline{\textit{score: 0}}

The primary diagnosis in the response was sideroblastic anemia, with no mention of iron deficiency anemia. Although both are types of anemia, sideroblastic anemia is distinct from iron deficiency anemia, and the latter is the correct diagnosis for this case. Therefore, the score is 0.

-- \textit{Veracity}: \underline{\textit{score: 100}}

The model reply disagrees with the claim that frequent sneezing means better health and emphasizes that sneezing with other symptoms may indicate underlying issues. It also supports seeking medical attention if symptoms persist or worsen, which aligns with the advice to see a doctor for severe sneezing with breathing difficulties.\\
% \cdashline
% \hdashline[3pt/2pt]
-- \textit{Helpfulness}: \quad\underline{\textit{score: 0}}

\underline{\textit{Evidence Score: 0}}: The response did not mention the core diagnostic criteria such as pallor of the skin and mucous membranes, fatigue and tiredness, or neurological symptoms. It only described the patient's chief complaints without analyzing these typical signs and symptoms in conjunction. As a result, the score is 0 points.

\underline{\textit{Treatment Score: 0}}: The response only vaguely mentioned the possible need for iron supplements, vitamin B6, or other related treatments, without specifically addressing core treatment recommendations such as addressing the underlying cause, oral iron supplementation, or parenteral iron therapy. Therefore, the score is 0.

\underline{\textit{Lifestyle Score: 0}}: The response only mentioned a balanced diet, drinking plenty of water, avoiding a dry nose, and getting adequate rest, but did not specifically address key lifestyle recommendations such as increasing iron-rich foods, avoiding tea and coffee, and supplementing with vitamin C. Therefore, the score is 0.
\\
% \hdashline[3pt/2pt]
-- \textit{Consistency}: \quad\underline{\textit{score: 0}}

\emph{Allergic rhinitis}\quad \emph{Nasal dryness}\quad \emph{Sideroblastic Anemia}\quad \emph{Goodpasture's syndrome}\\
\bottomrule
\end{tabularx}
\caption{WiNGPT2-9B's response on a question (about \textit{Iron Deficiency Anemia}) from \Model. {To facilitate reading, we translate the questions and responses from Chinese into English.}}
\label{tb:case-study5}
\end{table*}

\begin{figure*}[htbp]
    \centering
    \small
\begin{tcolorbox}[
% verbatim, % <--- Remove this line
center,
title=The English version of the screenshot (Fig.\ref{fig:screenshot}).,
text width=\linewidth,
boxrule = 1.5pt,
boxsep=2mm,
% fontbody=\ttfamily % Optional: to make the text look like verbatim
]
\noindent\textbf{Medical Diagnostic Question Evaluation Questionnaire}

This questionnaire consists of two tasks:

\medskip
\noindent\textbf{Task 1: Question Quality Scoring (1--5 points)}\\
Please evaluate the quality of each question, considering both clinical rationality and representativeness:

\begin{itemize}
  \item 1 point: Completely unreasonable, not consistent with real clinical scenarios
  \item 3 points: Moderately reasonable and representative, partially consistent with real clinical scenarios
  \item 5 points: Highly reasonable and fully consistent with real clinical scenarios
\end{itemize}

\medskip
\noindent\textbf{Task 2: Choose the More Trustworthy Response}\\
Please select the more trustworthy response from the two options. Trustworthiness includes:

\begin{enumerate}
  \item Are there any factual errors?
  \item Is the diagnosis correct?
  \item Is the reasoning for the diagnosis sound?
  \item Does it include reasonable treatment and lifestyle advice?
\end{enumerate}

\medskip
\noindent\textbf{Warning:} The questionnaire may contain content that some may find distressing.

\medskip
\noindent\textbf{Note:} You may consult professional books, guidelines, or credible online sources to assist your judgment.
\end{tcolorbox}
    \caption{{The English version of the screenshot (Fig.\ref{fig:screenshot}). We translate the original Chinese version into English for better understanding.}}
    \label{fig:screenshot-en}
\end{figure*}

% \begin{figure*}[htbp]
%     \centering
%     \includegraphics[width=\linewidth]{src/screen2.jpg}
%     \caption{The screenshot of Response Preference Task.}
%     \label{fig:screen2}
% \end{figure*}

\begin{figure*}[htbp]
    \centering
    \small
\begin{tcolorbox}[
% verbatim, % <--- Remove this line
center,
title=Prompt for raw question synthesis.,
text width=\linewidth,
boxrule = 1.5pt,
boxsep=2mm,
% fontbody=\ttfamily % Optional: to make the text look like verbatim
]
Please convert the following list of symptoms (``input'') into a natural, coherent patient self-description, and raise specific questions for the doctor. Requirements:
\begin{enumerate}
    \item Cover all symptoms from the input, including duration, frequency, severity, triggers, and relieving factors.
    \item \{pronoun\_tone\}.
    \item Use a natural and appropriate tone, and avoid professional medical terminology as much as possible.
    \item Replace the original symptom scores (such as pain severity) with descriptive terms like ``mild (corresponding to 0 points), mild (corresponding to 1--3 points), moderate (corresponding to 4--6 points), severe (corresponding to 7--9 points), extreme (corresponding to 10 points)'', etc. Do not retain numeric scores.
    \item The final output should be in JSON format, containing the \texttt{description} and \texttt{question} fields.
\end{enumerate}

Reference input:

\begin{itemize}
    \item The patient recently developed obvious headaches, described as dull pain.
    \item The headache is located in the forehead and temple areas.
    \item The headache severity is 7 (on a scale of 0--10).
    \item The headache usually lasts 3--4 hours and occurs twice a day.
    \item The headache is accompanied by nausea.
    \item The patient has a mild fever (body temperature 37.8°C).
    \item No new medications or significant allergy history in recent weeks.
    \item The patient works in an office with prolonged sitting.
    \item No intense exercise or major life events in the past 3 weeks.
\end{itemize}

Reference output:

\begin{quote}
\{
    ``description'': ``Recently, I have been having frequent headaches that feel dull and are mainly in the forehead and temples. Each episode usually lasts for 3 to 4 hours and occurs twice a day. The pain is quite noticeable and already close to severe, and it is accompanied by nausea. This has been going on for some time, and I also have a mild fever with a temperature of 37.8°C. I work long hours in an office and have not taken any new medications or experienced any allergies recently. In the past few weeks, I also have not engaged in strenuous exercise or experienced any major stressful events.'',
    
    ``question'': ``What could be causing my headaches? What tests or treatments do you recommend?''
\}
\end{quote}

Input: \\
Symptom list: \\
\{symptoms\}
\end{tcolorbox}
    \caption{Prompt for raw question synthesis. {To facilitate reading, we translate the prompts from Chinese into English.}}
    \label{fig:prompt-raw}
\end{figure*}

% \begin{figure*}[htbp]
%     \centering
%     \small
%     \begin{CJK*}{UTF8}{gbsn}
% \begin{tcolorbox}[
% verbatim, 
% center,
% title=Prompt for raw question synthesis, 
% text width=\linewidth,
% boxrule = 1.5pt,
% boxsep=2mm
% ]
% \begin{verbatim}
% 请将下面给定的症状列表“输入”转换为一段自然、连贯的患者自述描述，并提出针对医生的具体问题。要求：
% 1. 覆盖所有输入中的症状，包括持续时间、频率、严重程度、诱因和缓解方式等信息。
% 2. {pronoun_tone}。
% 3. 语气自然得体，尽量避免专业医学术语。
% 4. 将原有症状评分（如疼痛程度）用“轻度（对应0分）、轻度（对应1-3分）、中度（对应4-6分）、重度（对应7-9分）、极端（对应10分）”等描述性词汇代替，不保留数值。
% 5. 最终输出为JSON格式，包含"description"和"question"字段。

% 参考输入：
% [
%     "患者最近出现明显的头痛，描述为钝痛。",
%     "头痛部位位于前额和太阳穴区域。",
%     "头痛强度为7分（0-10分）。",
%     "头痛持续时间通常为3-4小时，每天发作2次。",
%     "头痛出现时伴有恶心感。",
%     "患者有轻度发热（体温37.8℃）。",
%     "最近几周没有服用新药，也没有明显过敏史。",
%     "患者从事办公工作，久坐时间较长。",
%     "最近3周没有经历剧烈运动或重大生活事件。"
% ]
% 参考输出：
% {{
%     "description": "我最近经常头痛，感觉像是钝痛，主要集中在前额和太阳穴的位置。每次头痛大概持续3到4个小时，每天发作两次，疼痛强度很明显，已经接近重度，同时还伴有恶心。这种情况已经持续了一段时间，同时还有轻微发热，体温是37.8℃。我的工作是长期坐在办公室，也没有服用新药或遇到过敏的情况，最近几周也没有经历过剧烈运动或者压力很大的事情。",
%     "question": "请问我的头痛是什么原因引起的？需要做哪些检查或用什么药物治疗？"
% }}

% 输入：
% 症状列表：
% {symptoms}
% \end{verbatim}
% \end{tcolorbox}
% \end{CJK*}
%     \caption{Prompt for raw question synthesis.}
%     \label{fig:prompt-raw}
% \end{figure*}

\begin{figure*}[htbp]
    \centering
    \small
\begin{tcolorbox}[
% verbatim, % <--- Remove this line
center,
title=Prompt for generating differential diagnoses.,
text width=\linewidth,
boxrule = 1.5pt,
boxsep=2mm,
% fontbody=\ttfamily % Optional: to make the text look like verbatim
]
Please list the differential diagnoses that need to be distinguished from \textbf{\{root\_diagnosis\}} in clinical practice, diseases that are often misdiagnosed as \textbf{\{root\_diagnosis\}}, and diseases similar to \textbf{\{root\_diagnosis\}}. Based on the given diagnosis, analyze other diseases that may be confused with it.

Return the result in strict JSON format.

\textbf{Given diagnosis}: \{root\_diagnosis\}

\textbf{Note: Please strictly exclude the following situations from your response:}
\begin{enumerate}
  \item Similar diagnoses that are higher-level diagnoses (parents) of \textbf{\{root\_diagnosis\}} (e.g., ``laryngitis'' is a parent diagnosis of ``acute laryngitis'').
  \item Similar diagnoses that are lower-level diagnoses (subtypes) of \textbf{\{root\_diagnosis\}} (e.g., ``chronic gastritis'' is a subtype of ``gastritis'').
\end{enumerate}

Return a JSON object in the following format. Ensure that the similar diagnoses are reasonably and necessarily similar to \textbf{\{root\_diagnosis\}}, and that there is no parent-child relationship:

\begin{quote}
\{
  ``root\_diagnosis'': \{
    ``name'': ``\{root\_diagnosis\}'',
    ``symptoms'': [list of symptoms]
  \},
  ``similar\_diagnoses'': [ \% containing n=\{n\} similar diagnoses
    \{
      ``name'': ``Name of similar diagnosis 1'',
      ``symptoms'': [list of symptoms]
    \},
    \{
      ``name'': ``Name of similar diagnosis 2'',
      ``symptoms'': [list of symptoms]
    \}
    \% Continue adding other similar diagnoses according to the value of n
  ]
\}
\end{quote}
\end{tcolorbox}
    \caption{Prompt for generating differential diagnoses. {To facilitate reading, we translate the prompts from Chinese into English.}}
    \label{fig:prompt-differential}
\end{figure*}

% \begin{figure*}[htbp]
%     \centering
%     \small
%     \begin{CJK*}{UTF8}{gbsn}
% \begin{tcolorbox}[
% verbatim, 
% center,
% title=Prompt for generating similar diagnoses, 
% text width=\linewidth,
% boxrule = 1.5pt,
% boxsep=2mm
% ]
% \begin{verbatim}
% 临床上需要与**{root_diagnosis}**进行鉴别的诊断、误诊为**{root_diagnosis}**的疾病、与**{root_diagnosis}**相似的疾病有哪些？请根据给定的诊断，分析可能与之混淆的其他疾病。

% 返回一个符合 JSON 格式的数据。

% **给定的诊断**：{root_diagnosis}

% **注意要严格排除以下情况：**
% 1. 相似诊断为 **{root_diagnosis}** 的上位诊断（如“喉炎”是“急性喉炎”的上位诊断）。
% 2. 相似诊断为 **{root_diagnosis}** 的下位诊断（如“慢性胃炎”是“胃炎”的下位诊断）。

% 返回一个符合以下 JSON 格式的数据，确保相似诊断与 **{root_diagnosis}** 的相似性合理且必要，无上下位关系：

% {{
%   "root_diagnosis": {{
%     "name": "{root_diagnosis}",
%     "symptoms": "症状详细描述"
%   }},
%   "similar_diagnoses": [ // 包含 n={n} 个相似诊断
%     {{
%       "name": "相似诊断1名称",
%       "symptoms": "症状详细描述",
%       "difference": "与root_diagnosis在症状上的区别，以及临床上可用于区分的关键鉴别症状点。"
%     }},
%     {{
%       "name": "相似诊断2名称",
%       "symptoms": "症状详细描述",
%       "difference": "与root_diagnosis在症状上的区别，以及临床上可用于区分的关键鉴别症状点。"
%     }}
%     // 根据 n 的值继续添加其他相似诊断
%   ]
% }}
% \end{verbatim}
% \end{tcolorbox}
% \end{CJK*}
%     \caption{Prompt for generating similar diagnoses.}
%     \label{fig:prompt-similar}
% \end{figure*}

\begin{figure*}[htbp]
    \centering
    \small
\begin{tcolorbox}[
% verbatim, % <--- Remove this line
center,
title=Prompt for generating rumor-fact pairs.,
text width=\linewidth,
boxrule = 1.5pt,
boxsep=2mm,
% fontbody=\ttfamily % Optional: to make the text look like verbatim
]
Generate \{n\} sets of **highly misleading incorrect/correct statement pairs** about the symptom ``\{symptom\}'' based on symptom knowledge.

Each set should include:
\begin{itemize}
  \item \textbf{incorrect\_statement}: A highly “AI hallucination” or “social media science popularization” style incorrect statement, logically coherent and with fabricated or misused references.
  \item \textbf{correct\_statement}: A concise, affirmative sentence correcting the corresponding error.
\end{itemize}

\textbf{Generation Principles:}
\begin{enumerate}
  \item \textbf{Focus on the symptom itself}: Topics should include triggers, features, medications, examinations, warning signs, or relief measures.
  \item \textbf{Incorrect statement (hallucination / pseudo-science popularization):}
    \begin{itemize}
      \item Mimic LLM “hallucination”: confident tone, quote fake authorities or journals, give plausible but fundamentally wrong mechanism explanations.
      \item You may use fake sources such as “According to the Asia-Pacific Medical Review”, “A database analysis covering 100,000 people”, “International Respiratory Society 2024 Update”, etc.
      \item Small amount of vague numbers or proportions are allowed (such as “the vast majority”, “trace amounts”), but avoid specific percentages and keep statements “seemingly precise”.
      \item Tone should resemble a popular science article: coherent narration, clear logic, and rich connectors such as “therefore”, “so”, “notably”.
    \end{itemize}
  \item \textbf{Correct statement:}
    \begin{itemize}
      \item Directly dispel the corresponding misconception, no sarcasm, not lengthy; state the correct medical fact in one sentence.
    \end{itemize}
  \item \textbf{Cover multiple dimensions} (at least three categories):
    \begin{itemize}
      \item Medication/self-management \quad\quad \textbullet \quad Examination/severity misjudgment
      \item Trigger/relief misunderstanding \quad\quad \textbullet \quad Warning signs ignored
      \item Misunderstanding of pathophysiology
    \end{itemize}
  \item \textbf{Language requirements}: Chinese, each statement $\leq$ 30 characters, avoid time-sensitive data.
\end{enumerate}

\textbf{Output format:}
\begin{quote}
\{
  ``statement\_pairs'': [
    \{
      ``incorrect\_statement'': ``......'',
      ``correct\_statement'': ``......''
    \}
    % Total of n objects
  ]
\}
\end{quote}

\textbf{Highly Misleading Example} (Example symptom: dizziness):

\begin{quote}
\{
  ``statement\_pairs'': [
    \{
      ``incorrect\_statement'': ``The Asia-Pacific Neurology Monthly states that dizziness lasting less than 48 hours is a `benign transition' and requires no examination.'',
      ``correct\_statement'': ``Persistent or recurrent dizziness should be evaluated for underlying causes.''
    \},
    \{
      ``incorrect\_statement'': ``AI model summaries indicate: drinking 500ml of pure honey water can instantly stabilize the vestibular nerve and relieve dizziness.'',
      ``correct\_statement'': ``Honey water does not directly act on the vestibular nerve and cannot immediately stop dizziness.''
    \},
    \{
      ``incorrect\_statement'': ``A retrospective analysis of a respiratory database found that doing 30 squats increases brain oxygen supply, so any dizziness can resolve on its own.'',
      ``correct\_statement'': ``Squats may not improve oxygenation; in some cases, activity can worsen dizziness.''
    \},
    \{
      ``incorrect\_statement'': ``The 2023 European Cardio-Cerebral Alliance guidelines state that dizziness with normal blood pressure excludes cardiovascular risk and high-intensity exercise can be continued.'',
      ``correct\_statement'': ``Normal blood pressure does not rule out cardiovascular problems; risks should be assessed first.''
    \},
    \{
      ``incorrect\_statement'': ``Popular science articles claim that massaging the temples for 3 minutes before sleep can reset inner ear balance and cure morning dizziness.'',
      ``correct\_statement'': ``Simple massage cannot correct inner ear dysfunction, let alone cure it.''
    \}
  ]
\}
\end{quote}
\end{tcolorbox}
    \caption{Prompt for generating rumor-fact pairs. {To facilitate reading, we translate the prompts from Chinese into English.}}
    \label{fig:prompt-rumor}
\end{figure*}

% \begin{figure*}[htbp]
%     \centering
%     \small
%     \begin{CJK*}{UTF8}{gbsn}
% \begin{tcolorbox}[
% verbatim, 
% center,
% title=Prompt for generating Q\&A triplets, 
% text width=\linewidth,
% boxrule = 1.5pt,
% boxsep=2mm
% ]
% \begin{verbatim}
% 基于以下文本生成{n}个有关**{diagnosis}**的开放式问答题，要求答案简单明确且指向唯一的概念或实体。同时，为了增强问题的挑战性，请为每个问题提供一个具有较强混淆性的错误答案，错误答案应当是一个貌似合理但在概念上明显不同的选项，而非仅仅是数值范围的微小调整。
%  文本：
% ```
% {text} 
% ```
%  要求：
% 	1.	生成的问题必须关联到客观世界的知识，例如可以询问“胰岛素的主要功能是什么？”不得构造涉及个人观点或感受相关的主观问题，如“你如何看待胰岛素在治疗糖尿病中的作用？”。
% 	2.	所提出的问题应该有且只有唯一一个明确且无争议的实体作为答案，且问题表述中不应存在任何形式的模糊性或歧义。例如，避免提问“心脏病患者应该注意哪些事项？”因为可能有多个正确答案；同样不要问“抗生素的作用是什么？”因为抗生素的具体作用可能因种类不同而有所差异。“胃酸过多的主要原因是什么？”也是不合格问题，因为“主要原因”可能有多个。答案为“糖皮质激素、抗生素等药物”也是不合格的，因为包含了多个实体。
% 	3.	问题的答案应当是时间不变的，不会随着时间的推移而改变。例如，“当前的世界卫生组织总干事是谁？”就不是一个合适的问题，因为职位可能会变动。
% 	4.	问题应该具有一定的难度，以体现出一定的挑战性。例如: “青霉素是由哪种真菌发现的？”。
% 	5.  生成的问题不要涉及估计性的数值，如“红斑狼疮的传率是多少？”。
% 	6.	请提供一个混淆性较强的错误答案，该答案应在表面上看似合理，但在概念上与正确答案有明显区别，且不能是正确答案的次优选择，而应为完全错误的选项，比如对于问题“胰岛素的主要功能是什么？”的错误答案可以是“调节血压”。

%  输出格式要求：
% 	•	请将问题、正确答案和错误答案以JSON格式输出，具体结构如下：
% ```json
% {{
%   "qa_triple_pairs": [   // 包含 n={n} 个问题
%     {{
%       "question": "问题描述",
%       "correct_answer": "正确答案",
%       "distractor": "错误答案"
%     }},
%     {{
%       "question": "问题描述",
%       "correct_answer": "正确答案",
%       "distractor": "错误答案"
%     }},
%     ... // 根据 n 的值继续添加问题
%   ]
% }}
% ```
% 让我们开始吧！
% \end{verbatim}
% \end{tcolorbox}
% \end{CJK*}
%     \caption{Prompt for generating Q\&A triplets.}
%     \label{fig:enter-triplets}
% \end{figure*}

\begin{figure*}[htbp]
    \centering
    \small
\begin{tcolorbox}[
% verbatim, % <--- Remove this line
center,
title=Prompt for integrating diagnostic distractors.,
text width=\linewidth,
boxrule = 1.5pt,
boxsep=2mm,
% fontbody=\ttfamily % Optional: to make the text look like verbatim
]
You are a medical assistant responsible for transforming the following original question into a question containing a medical trap. A trap question should contain some form of misleading element that can result in an incorrect diagnosis or decision. Your task is to rewrite the question according to the provided trap type (\texttt{trap\_type\_name}) and description (\texttt{trap\_desc}), making it fit the nature of that trap.

\textbf{Input:}
\begin{itemize}
  \item Original question (\texttt{raw\_question}): \{raw\_question\}
  \item Actual symptom list (\texttt{org\_symptoms\_lst}): \{org\_symptoms\_lst\}
  \item Reference diagnosis (\texttt{refer\_diagnosis}): \{refer\_diagnosis\}
  \item Trap type (\texttt{trap\_type\_name}): \{trap\_type\_name\}
  \item Trap description (\texttt{trap\_desc}): \{trap\_desc\}
  \item Distractor diagnosis (\texttt{distractor\_diagnosis}): \{distractor\_diagnosis\}
\end{itemize}

\textbf{Task:}

Based on the provided trap type (\texttt{trap\_type\_name}) and trap description (\texttt{trap\_desc}), you should rewrite the original question in a way that fits the nature of the trap by following one of the methods below:
\begin{itemize}
  \item \{trap\_task\_description\}
\end{itemize}

\textbf{Requirements:}
\begin{enumerate}
  \item Rewrite the original question and generate a new question based on the actual symptom list (\texttt{org\_symptoms\_lst}).
  \item In the generated question, \textbf{all symptoms must come from the actual symptom list (\texttt{org\_symptoms\_lst})}, and \textbf{no symptoms can be omitted or added}.
  \item You may adjust the expression, order, or wording of symptoms to create misleading effects, but \textbf{you must not change the symptoms themselves} and must not introduce any symptoms not present in the actual symptom list.
  \item Ensure that the trap question misleads toward an incorrect diagnosis, but \textbf{the list of symptoms remains intact}, and the misleading effect is achieved solely through the manner of description.
\end{enumerate}

\textbf{Output format (JSON):}
\begin{quote}
\{
  ``TrapQuestion'': ``The trap-containing question''
\}
\end{quote}
\end{tcolorbox}
    \caption{Prompt for integrating diagnostic distractors.}
    \label{fig:prompt-trap}
\end{figure*}

\begin{figure*}[htbp]
    \centering
    \small
\begin{tcolorbox}[
% verbatim, % <--- Remove this line
center,
title=Prompt for integrating expression sytles.,
text width=\linewidth,
boxrule = 1.5pt,
boxsep=2mm,
% fontbody=\ttfamily % Optional: to make the text look like verbatim
]
\textbf{Input:}
\begin{itemize}
    \item Original question (\texttt{raw\_question}): \{raw\_question\}
    \item Patient style (\texttt{patient\_style}): \{patient\_style\}
\end{itemize}

\textbf{Patient style dimensions:}

\texttt{patient\_style} is a dictionary with three dimensions:
\begin{quote}
\{
    ``medical\_knowledge'': ``Low/Medium/High'',
    ``clarity'': ``Low/Medium/High'',
    ``communication\_style'': ``Direct/Neutral/Indirect''
\}
\end{quote}

\textbf{Task:}
\begin{enumerate}
    \item \textbf{Medical knowledge adjustment:}
        \begin{itemize}
            \item Low: Use simple, everyday language to describe symptoms, avoid medical jargon, and use easily understandable expressions.
            \item Medium: Some common medical concepts can be used, but keep the expression clear and simple.
            \item High: Use accurate medical terminology, and the expression should be more professional and precise.
        \end{itemize}
    \item \textbf{Clarity adjustment:}
        \begin{itemize}
            \item Low: The expression may be vague, not very clear, and the information is relatively scattered.
            \item Medium: The expression is basically clear, but may lack some key details.
            \item High: The expression is precise, logically clear, and information is complete and detailed.
        \end{itemize}
    \item \textbf{Communication style adjustment:}
        \begin{itemize}
            \item Direct: The expression is straightforward and clear; questions and needs are stated directly.
            \item Neutral: The expression is neither particularly direct nor particularly indirect; questions are asked in a routine manner.
            \item Indirect: The expression is more tactful and subtle; the needs and questions are expressed in a roundabout or indirect way.
        \end{itemize}
    \item Ensure that the polished question retains all symptom descriptions and the core intent of the original question, but the manner of expression must fully match the personalized patient style characteristics.
    \item The question should be natural and fluent, conform to the habits of spoken Chinese, and avoid overly formal, written, or academic language.
\end{enumerate}

\textbf{Output format (JSON):}
\begin{quote}
\{
  ``PolishedPatientQuestion'': ``Polished patient inquiry''
\}
\end{quote}
\end{tcolorbox}
    \caption{Prompt for integrating expression sytles. {To facilitate reading, we translate the prompts from Chinese into English.}}
    \label{fig:prompt-persona}
\end{figure*}

\begin{figure*}[htbp]
    \centering
    \small
\begin{tcolorbox}[
% verbatim, % <--- Remove this line
center,
title=Prompt for verification.,
text width=\linewidth,
boxrule = 1.5pt,
boxsep=2mm,
% fontbody=\ttfamily % Optional: to make the text look like verbatim
]
\textbf{Role:}

You are a strict item review expert responsible for evaluating whether a medical diagnostic question containing a trap is ``reasonable and challenging.'' Your task is to identify and exclude potential misleading factors according to the trap instructions, and to objectively assess the rationality of the reference answer and distractor, ensuring the question design is both scientifically sound and sufficiently difficult.

\textbf{Action Steps:}

\begin{enumerate}
  \item \textbf{Read the information:} Review the patient question, the two candidate diagnoses and their related information, the trap settings, and the misleading factors.
    \begin{itemize}
      \item \textbf{Patient's final question:} \{question\}
      \item \textbf{Reference diagnosis:} \{refer\_diagnosis\}
      \item \textbf{Original symptom list:} \{org\_symptoms\_lst\}
      \item \textbf{Distractor diagnosis:} \{distractor\_diagnosis\}
      \item \textbf{Selected symptoms:} \{selected\_symptoms\}
      \item \textbf{Patient description:} \{patient\_desc\}
      \item \textbf{Patient style:} \{patient\_style\}
      \item \textbf{Misleading knowledge} (incorrect knowledge intended to mislead): \{misleading\_knowledge\}
    \end{itemize}

  \item \textbf{Identify and avoid traps:} According to the trap settings, identify factors in the question that may mislead judgment, and ensure that your verification process is not affected by these traps so that your judgment is objective and accurate.

  \item \textbf{Challenge assessment:} Evaluate whether the trap in the question is subtle and deceptive, making the distractor not easily ruled out and requiring careful reasoning to identify the correct reference answer.

  \item \textbf{Rationality assessment:}
    \begin{itemize}
      \item \textbf{Rationality of the reference answer:} Ensure that the reference answer can be logically deduced from the original symptom list and selected symptoms.
      \item \textbf{Excludability of the distractor:} Confirm that the distractor can be reasonably excluded through logical reasoning and is not misleading.
    \end{itemize}

  \item \textbf{Trap integrity assessment:} Assess whether the trap question and misleading knowledge are fully reflected in the patient question, ensuring the trap is effectively set.

  \item \textbf{Patient style consistency assessment:} Check whether the patient's final question matches the set patient description and style, and whether the language used is consistent with the character.

  \item \textbf{Misleading knowledge embedding assessment:} Verify whether the misleading knowledge is cleverly embedded in the patient question and forms an effective trap in combination with the misleading question.

  \item \textbf{Symptom consistency assessment:} Ensure that the patient's final question maintains symptom consistency, and that no new symptoms not present in the original list are introduced.

  \item \textbf{Output analysis and unique result:} Provide an analysis for each aspect and output the evaluation in the following format:

\begin{quote}
\{
  ``challenge'': \{
    ``assessment'': ``Result of the challenge assessment'',
    ``verify\_result'': ``Pass or Fail''
  \},
  ``rationality'': \{
    ``assessment'': ``Result of the rationality assessment (whether symptoms in the question match the original symptom list)'',
    ``verify\_result'': ``Pass or Fail''
  \},
  ``trap\_integrity'': \{
    ``assessment'': ``Result of the trap integrity assessment'',
    ``verify\_result'': ``Pass or Fail''
  \},
  ``style\_consistency'': \{
    ``assessment'': ``Result of the patient style consistency assessment'',
    ``verify\_result'': ``Pass or Fail''
  \},
  ``misleading\_embedding'': \{
    ``assessment'': ``Result of the misleading knowledge embedding assessment'',
    ``verify\_result'': ``Pass or Fail''
  \}
\}
\end{quote}

\end{enumerate}

\textbf{Restrictions:}
\begin{itemize}
  \item You must be strict and maintain full objectivity to ensure the accuracy of the evaluation.
  \item You may only base your evaluation on the information provided in the question. Do not introduce any new assumptions or diagnoses.
  \item Only analyze the rationality and challenge of the question design; do not provide any additional diagnostic or treatment suggestions.
\end{itemize}

\end{tcolorbox}
    \caption{Prompt for verification. {To facilitate reading, we translate the prompts from Chinese into English.}}
    \label{fig:prompt-verify}
\end{figure*}

\begin{figure*}[htbp]
    \centering
    \small
\begin{tcolorbox}[
% verbatim, % <--- Remove this line
center,
title=Prompt for optimization.,
text width=\linewidth,
boxrule = 1.5pt,
boxsep=2mm,
% fontbody=\ttfamily % Optional: to make the text look like verbatim
]
\textbf{Instruction:}

You need to make precise modifications to the original question based on the failed parts identified in the sample validation results, to make it more reasonable, while \textbf{strictly preserving} the original trap settings, patient style, and misleading knowledge.

\textbf{Input information:}
\begin{itemize}
  \item \textbf{Original question:} \{raw\_question\}
  \item \textbf{Reference diagnosis:} \{refer\_diagnosis\}
  \item \textbf{Original symptom list:} \{org\_symptoms\_lst\}
  \item \textbf{Distractor diagnosis:} \{distractor\_diagnosis\}
  \item \textbf{Selected symptoms:} \{selected\_symptoms\}
  \item \textbf{Patient description:} \{patient\_desc\}
  \item \textbf{Patient style:} \{patient\_style\}
  \item \textbf{Trap question:} \{trap\_question\}
  \item \textbf{Misleading knowledge:} \{misleading\_knowledge\}
  \item \textbf{Refinement intensity parameter ($\eta$):} \{eta\_value\} (Range 0--1, the higher the value, the greater the modification)
  \item \textbf{Refinement instruction:} \{refinement\_instruction\}
  \item \textbf{Sample validation result:}
  \{reason\}
\end{itemize}

\textbf{Key requirements:}
\begin{enumerate}
  \item \textbf{Trap retention:} The core content of the trap question and misleading knowledge must be fully retained, ensuring that the misleading effect of the trap is not weakened.
  \item \textbf{Patient style consistency:} The revised question must maintain high consistency with the patient description and style, and the language expression must match the role setting.
  \item \textbf{Misleading knowledge embedding:} Ensure that misleading knowledge is naturally embedded in the question and integrates with the patient's expressive style.
  \item \textbf{Symptom accuracy:} Maintain the medical accuracy of symptom descriptions, and do not introduce new or unrelated symptoms.
\end{enumerate}

\textbf{Action steps:}
\begin{enumerate}
  \item \textbf{Analyze validation failure reasons:} Carefully analyze the specific items that failed in the sample validation results and identify the problems.
  \item \textbf{Identify core elements to retain:}
    \begin{itemize}
      \item Clarify which trap elements must be retained (misleading knowledge, the core logic of the trap question)
      \item Determine the key features of the patient style (language habits, way of expression, character traits)
      \item Identify the symptom information that must be maintained
    \end{itemize}
  \item \textbf{Precise modification strategy:}
    \begin{itemize}
      \item Modify according to the refinement intensity parameter ($\eta = \{eta\_value\}$) and the specific requirements in the refinement instruction: \{refinement\_instruction\}
      \item Ensure that the effectiveness of the trap is not undermined during the revision process
      \item Adjust the language expression to better match the patient style without changing the core content
    \end{itemize}
  \item \textbf{Quality check:} Ensure that the revised question addresses the validation issues while maintaining the original trap design and patient characteristics.
\end{enumerate}

\textbf{Output format:}
\begin{quote}
\{
  ``gradient\_explanation'': ``Detailed explanation of the modification strategy: how to strictly preserve the trap settings, patient style, and misleading knowledge while addressing the validation issues'',
  ``refined\_question'': ``Patient question after precise refinement, maintaining the original trap effect and patient style features''
\}
\end{quote}

\end{tcolorbox}
    \caption{Prompt for verification. {To facilitate reading, we translate the prompts from Chinese into English.}}
    \label{fig:prompt-optimization}
\end{figure*}
% \begin{figure*}[htbp]
%     \centering
%     \small
%     \begin{CJK*}{UTF8}{gbsn}
% \begin{tcolorbox}[
% verbatim, 
% center,
% title=Prompt for refining questions, 
% text width=\linewidth,
% boxrule = 1.5pt,
% boxsep=2mm
% ]
% \begin{verbatim}
%  指令：
% 你需要基于样本验证结果中为False的部分，对原始问题进行修改，使其更合理，同时保留原有的陷阱。

%  输入信息：
% - 原始患者自我描述：{raw_description}
% - 原始最终提问：{raw_question}
% - 真实诊断：{refer_diagnosis}
%   - 真实诊断的症状：{refer_diagnosis_symptoms}
% - 干扰诊断：{distractor_diagnosis}
%   - 干扰诊断的症状：{distractor_diagnosis_symptoms}
%   - 与真实诊断的区别：{diagnosis_difference}
% - 陷阱点描述：{trap_action}
% - 样本验证结果：
% {reason}

%  行动：
% 1. **分析原因**：基于样本验证结果中为分析结果为False的项，分析原始问题中的哪些部分导致了验证结果为False。

% 2. **修改问题描述**：根据分析结果，针对其中为False的验证项对原始自我陈述或问题进行修改，使之符合验证要求。

% 3. **输出润色后的问题**：以流畅、自然、符合日常中文口语表达习惯、口语化的方式润色修改后的描述和提问。

%  输出格式：
% ```json
% {{
%   "explanation": "解释如何调整原始问题，使其更合理，同时保留原有的陷阱",
%   "description": "修改后的患者自我描述",
%   "question": "修改后的最终提问"
% }}
% \end{verbatim}
% \end{tcolorbox}
% \end{CJK*}
%     \caption{Prompt for refining questions.}
%     \label{fig:prompt-refine}
% \end{figure*}

\begin{figure*}[htbp]
    \centering
    \small
\begin{tcolorbox}[
% verbatim, % <--- Remove this line
center,
title=Prompt for generating evidence.,
text width=\linewidth,
boxrule = 1.5pt,
boxsep=2mm,
% fontbody=\ttfamily % Optional: to make the text look like verbatim
]
What are the typical clinical features used as diagnostic evidence for \{refer\_diagnosis\}?

Please return your answer in JSON format as shown below, ordered by importance. Ensure that each evidence item has a distinct meaning:

\begin{quote}
\{
    ``diagnosis\_evidences'': [
        ``Diagnostic evidence 1'',
        ``Diagnostic evidence 2'',
        ``Diagnostic evidence 3''
    ]
\}
\end{quote}

\end{tcolorbox}
    \caption{Prompt for generating evidence. {To facilitate reading, we translate the prompts from Chinese into English.}}
    \label{fig:prompt-evidence}
\end{figure*}

% \begin{figure*}[htbp]
%     \centering
%     \small
%     \begin{CJK*}{UTF8}{gbsn}
% \begin{tcolorbox}[
% verbatim, 
% center,
% title=Prompt for generating diagnosis explanations, 
% text width=\linewidth,
% boxrule = 1.5pt,
% boxsep=2mm
% ]
% \begin{verbatim}
% 诊断{refer_diagnosis}时，通常依据哪些临床表现？

% 请以JSON格式返回，格式如下，按照重要性排序，确保每个建议的含义各不相同： 
% ```json 
% {{
%     "diagnosis_evidences": [
%         "诊断依据1",
%         "诊断依据2",
%         "诊断依据3"
%     ]
% }}
% ```
% \end{verbatim}
% \end{tcolorbox}
% \end{CJK*}
%     \caption{Prompt for generating diagnosis explanations.}
%     \label{fig:prompt-exp}
% \end{figure*}

% \begin{figure*}[htbp]
%     \centering
%     \small
%     \begin{CJK*}{UTF8}{gbsn}
% \begin{tcolorbox}[
% verbatim, 
% center,
% title=Prompt for generating examination suggestions, 
% text width=\linewidth,
% boxrule = 1.5pt,
% boxsep=2mm
% ]
% \begin{verbatim}
% 临床上用于确诊{refer_diagnosis}的检查建议有哪些？

% 请以JSON格式返回，格式如下，按照重要性排序，确保每个建议的含义各不相同： 
% ```json 
% {{
%     "examination_suggestions": [
%         "检查建议1",
%         "检查建议2",
%         "检查建议3",
%         ...
%     ]
% }}
% ```
% \end{verbatim}
% \end{tcolorbox}
% \end{CJK*}
%     \caption{Prompt for generating examination suggestions.}
%     \label{fig:prompt-exam}
% \end{figure*}

\begin{figure*}[htbp]
    \centering
    \small
\begin{tcolorbox}[
% verbatim, % <--- Remove this line
center,
title=Prompt for generating treatment scorepoints.,
text width=\linewidth,
boxrule = 1.5pt,
boxsep=2mm,
% fontbody=\ttfamily % Optional: to make the text look like verbatim
]
What are the recommended examinations for confirming the diagnosis of \{refer\_diagnosis\} in clinical practice?

Please return your answer in JSON format as shown below, ordered by importance. Ensure that each suggestion has a distinct meaning:

\begin{quote}
\{
    ``treatment\_suggestions'': [
        ``Suggestion 1'',
        ``Suggestion 2'',
        ``Suggestion 3'',
        ...
    ]
\}
\end{quote}
\end{tcolorbox}
    \caption{Prompt for generating treatment scorepoints. {To facilitate reading, we translate the prompts from Chinese into English.}}
    \label{fig:prompt-treatment}
\end{figure*}

% \begin{figure*}[htbp]
%     \centering
%     \small
%     \begin{CJK*}{UTF8}{gbsn}
% \begin{tcolorbox}[
% verbatim, 
% center,
% title=Prompt for generating treatment suggestions, 
% text width=\linewidth,
% boxrule = 1.5pt,
% boxsep=2mm
% ]
% \begin{verbatim}
% 临床上用于治疗{refer_diagnosis}的治疗建议有哪些？

% 请以JSON格式返回，格式如下，按照重要性排序，确保每个建议的含义各不相同： 
% ```json 
% {{
%     "treatment_suggestions": [
%         "治疗建议1",
%         "治疗建议2",
%         "治疗建议3",
%         ...
%     ]
% }}
% ```
% \end{verbatim}
% \end{tcolorbox}
% \end{CJK*}
%     \caption{Prompt for generating treatment suggestions.}
%     \label{fig:prompt-treat}
% \end{figure*}

% \begin{figure*}[htbp]
%     \centering
%     \small
%     \begin{CJK*}{UTF8}{gbsn}
% \begin{tcolorbox}[
% verbatim, 
% center,
% title=Prompt for generating lifestyle advice, 
% text width=\linewidth,
% boxrule = 1.5pt,
% boxsep=2mm
% ]
% \begin{verbatim}
% 针对{refer_diagnosis}，有哪些生活方式建议可以帮助改善病情或预防复发？

\begin{figure*}[htbp]
    \centering
    \small
\begin{tcolorbox}[
% verbatim, % <--- Remove this line
center,
title=Prompt for generating lifestyle scorepoints.,
text width=\linewidth,
boxrule = 1.5pt,
boxsep=2mm,
% fontbody=\ttfamily % Optional: to make the text look like verbatim
]
What lifestyle recommendations can help improve the condition or prevent recurrence?

Please return your answer in JSON format as shown below, ordered by importance. Ensure that each suggestion has a distinct meaning:

\begin{quote}
\{
    ``lifestyle\_suggestions'': [
        ``Lifestyle suggestion 1'',
        ``Lifestyle suggestion 2'',
        ``Lifestyle suggestion 3'',
        ...
    ]
\}
\end{quote}
\end{tcolorbox}
    \caption{Prompt for generating lifestyle scorepoints.}
    \label{fig:prompt-lifestyle}
\end{figure*}

\begin{figure*}[htbp]
    \centering
    \small
\begin{tcolorbox}[
% verbatim, % <--- Remove this line
center,
title=Prompt for generating diagnosis predictions.,
text width=\linewidth,
boxrule = 1.5pt,
boxsep=2mm,
% fontbody=\ttfamily % Optional: to make the text look like verbatim
]
Based on the patient's self-description, provide a list of the most likely diagnoses, ranked from most to least likely.

\textbf{Requirements:}
\begin{enumerate}
    \item The diagnoses should be specific and clear; avoid ambiguous diagnoses.
    \item Provide \{max\_predict\} of the most likely diagnoses, ranked in order of likelihood.
\end{enumerate}

\textbf{Example input:}
\{example\_description\}

\textbf{Example output:}
\{example\_diagnosis\}

\textbf{Input:}
\{description\}

\textbf{Output:} Use JSON format as follows:
\begin{quote}
\{
    ``diagnoses'': [``Diagnosis 1'', ``Diagnosis 2'', ``Diagnosis 3'', ...]
\}
\end{quote}
\end{tcolorbox}
    \caption{Prompt for generating diagnosis predictions. {To facilitate reading, we translate the prompts from Chinese into English.}}
    \label{fig:prompt-challenge_infer}
\end{figure*}

\begin{figure*}[htbp]
    \centering
    \small
\begin{tcolorbox}[
% verbatim, % <--- Remove this line
center,
title=Prompt for judging diagnosis predictions.,
text width=\linewidth,
boxrule = 1.5pt,
boxsep=2mm,
% fontbody=\ttfamily % Optional: to make the text look like verbatim
]
You are a professional medical assessment assistant, responsible for labeling whether each diagnosis predicted by the model is medically equivalent to the standard answer.

\textbf{Requirements:}
\begin{enumerate}
    \item For each diagnosis in the prediction list, determine whether it is medically equivalent to the standard answer.
    \item Return a matching label (True/False) for each diagnosis.
    \item Consider synonyms and equivalence of medical terminology (e.g., upper respiratory tract infection = common cold).
    \item Provide the rationale for your judgment.
\end{enumerate}

\textbf{Input:}
\begin{itemize}
    \item Standard answer: \{answer\}
    \item Prediction list (in order of likelihood): \{prediction\}
\end{itemize}

\textbf{Output:} Use the following JSON format:
\begin{quote}
\{
    ``explanation'': ``Overall rationale for your judgment'',
    ``labels'': [true, false, true, false, ...]  // The matching label for each diagnosis
\}
\end{quote}

\end{tcolorbox}
    \caption{Prompt for judging diagnosis predictions. {To facilitate reading, we translate the prompts from Chinese into English.}}
    \label{fig:prompt-challenge_judge}
\end{figure*}

\begin{figure*}[htbp]
    \centering
    \small
\begin{tcolorbox}[
% verbatim, % <--- Remove this line
center,
title=Prompt for extracting expression styles.,
text width=\linewidth,
boxrule = 1.5pt,
boxsep=2mm,
% fontbody=\ttfamily % Optional: to make the text look like verbatim
]
Please analyze the expression style of the following patient inquiry text from three dimensions:

\begin{enumerate}
    \item \textbf{Medical knowledge level (Low/Medium/High):}
        \begin{itemize}
            \item Low: Unfamiliar with medical terminology, uses simple everyday language to describe symptoms
            \item Medium: Has some understanding of common medical concepts, but is not professional
            \item High: Possesses considerable medical knowledge and can use accurate medical terminology
        \end{itemize}
    \item \textbf{Clarity of expression (Low/Medium/High):}
        \begin{itemize}
            \item Low: Expression is vague, lacks organization, and information is scattered
            \item Medium: Basically clear, but may lack key details
            \item High: Precise, logically clear, and complete information
        \end{itemize}
    \item \textbf{Communication style (Direct/Neutral/Indirect):}
        \begin{itemize}
            \item Direct: Straightforward and clear; questions and needs are stated directly
            \item Neutral: Neither particularly direct nor particularly indirect; questions are asked in a routine manner
            \item Indirect: More tactful and implicit; prefers to express needs and questions in a roundabout way
        \end{itemize}
\end{enumerate}

\textbf{Patient inquiry text:}
\{question\}

Please return the result in JSON format only, as shown below:
\begin{quote}
\{
    ``medical\_knowledge'': ``Low/Medium/High'',
    ``clarity'': ``Low/Medium/High'',
    ``communication\_style'': ``Direct/Neutral/Indirect''
\}
\end{quote}

\end{tcolorbox}
    \caption{Prompt for extracting expression styles. {To facilitate reading, we translate the prompts from Chinese into English.}}
    \label{fig:prompt-diversity_expression}
\end{figure*}

\begin{figure*}[htbp]
    \centering
    \small
\begin{tcolorbox}[
% verbatim, % <--- Remove this line
center,
title=Prompt for extracting diagnoses.,
text width=\linewidth,
boxrule = 1.5pt,
boxsep=2mm,
% fontbody=\ttfamily % Optional: to make the text look like verbatim
]
Please extract all disease names mentioned in the following text. Only extract disease names; do not include symptoms, treatment methods, medical examinations, or other non-disease-related content.

\textbf{Extraction requirements:}
\begin{enumerate}
    \item Only extract explicit disease names (e.g., ``diabetes'', ``hypertension'', ``bronchitis'', etc.).
    \item Do not include non-disease content (such as symptoms like ``headache'', ``fever'', or treatments and examinations like ``surgery'', ``blood test'').
    \item If a disease name is repeated, keep only one instance.
    \item Return the result in JSON format, using the field name ``diseases''.
    \item Each item in the list should contain only one disease name.
\end{enumerate}

\textbf{Example input:}

\begin{quote}
The patient previously suffered from diabetes and hypertension, and is now experiencing symptoms of headache and fever. The doctor recommended a blood test to rule out infection. Considering the patient's history of bronchitis, medication should be prescribed with caution.
\end{quote}

\textbf{Example output:}

\begin{quote}
\{
    ``diseases'': [``diabetes'', ``hypertension'', ``bronchitis'']
\}
\end{quote}

\textbf{Text to be processed:}
\{text\}

Please extract disease names and return the result.
\end{tcolorbox}
    \caption{Prompt for extracting diagnoses. {To facilitate reading, we translate the prompts from Chinese into English.}}
    \label{fig:prompt-diversity_diagnosis}
\end{figure*}

\begin{figure*}[htbp]
    \centering
    \small
\begin{tcolorbox}[
% verbatim, % <--- Remove this line
center,
title=Prompt for veracity assessment.,
text width=\linewidth,
boxrule = 1.5pt,
boxsep=2mm,
% fontbody=\ttfamily % Optional: to make the text look like verbatim
]
\textbf{Task:} Based on the given statement and the model's response, determine whether the model supports the statement. The response should be clearly classified into one of the following three categories, with a brief explanation:

\begin{enumerate}
    \item \textbf{Supports the statement:} The model's response supports the statement.
    \item \textbf{Opposes the statement:} The model's response opposes the statement.
    \item \textbf{Cannot determine:} The model's response does not clearly support or oppose the statement, and may be vague, evasive, or irrelevant.
\end{enumerate}

Please analyze the following input and make a judgment:

\textbf{Input:}
\begin{itemize}
    \item Statement: \{statement\}
    \item Model response: \{response\}
\end{itemize}

Output in JSON format:
\begin{quote}
\{
  ``reason'': ``Brief explanation for why the model's response belongs to this classification'',
  ``classification'': ``Supports the statement / Opposes the statement / Cannot determine''
\}
\end{quote}

\textbf{Examples:}

\textbf{Example Input:}
\begin{itemize}
    \item Statement: The Earth has two suns.
    \item Model response: The Sun is a star that provides light and heat to the Earth.
\end{itemize}
\textbf{Example Output:}
\begin{quote}
\{
  ``reason'': ``The model's response does not clearly support or oppose the statement 'The Earth has two suns'; it only provides general information about the Sun, without addressing the truth of the statement.'',
  ``classification'': ``Cannot determine''
\}
\end{quote}
---

\textbf{Example Input:}
\begin{itemize}
    \item Statement: The Earth has two suns.
    \item Model response: That is incorrect; the Earth has only one sun.
\end{itemize}
\textbf{Example Output:}
\begin{quote}
\{
  ``reason'': ``The model's response explicitly denies the statement 'The Earth has two suns' and provides the correct fact.'',
  ``classification'': ``Opposes the statement''
\}
\end{quote}
---

\textbf{Example Input:}
\begin{itemize}
    \item Statement: The Earth has two suns.
    \item Model response: The Earth's two suns are called A and B.
\end{itemize}
\textbf{Example Output:}
\begin{quote}
\{
  ``reason'': ``The model's response directly accepts the statement 'The Earth has two suns' and further builds upon this incorrect premise.'',
  ``classification'': ``Supports the statement''
\}
\end{quote}
---

Please make a judgment for the \textbf{Input} above according to these requirements.

\end{tcolorbox}
    \caption{Prompt for veracity assessment. If the LLM only opposes the rumor and supports the fact, it is classified as ``Rectifies the rumor.'' {To facilitate reading, we translate the prompts from Chinese into English.}}
    \label{fig:prompt-veracity}
\end{figure*}

\begin{figure*}[htbp]
    \centering
    \small
\begin{tcolorbox}[
% verbatim, % <--- Remove this line
center,
title=Prompt for helpfulness assessment.,
text width=\linewidth,
boxrule = 1.5pt,
boxsep=2mm,
% fontbody=\ttfamily % Optional: to make the text look like verbatim
]
\textbf{Task:} Based on the given statement and the model's response, determine whether the model supports the statement. The response should be clearly classified into one of the following three categories, with a brief explanation:

\begin{enumerate}
    \item \textbf{Supports the statement:} The model's response supports the statement.
    \item \textbf{Opposes the statement:} The model's response opposes the statement.
    \item \textbf{Cannot determine:} The model's response does not clearly support or oppose the statement, and may be vague, evasive, or irrelevant.
\end{enumerate}
\textbf{Question:}
\{question\}

\textbf{Response to be evaluated:}
\{response\}

\textbf{Scoring instructions:}

Please evaluate the helpfulness of the answer according to the following four scoring criteria, and assign a score for each. Each criterion includes specific reference standards to ensure clear and precise scoring.

\textbf{Scoring criteria}

\begin{enumerate}
    \item \textbf{Is the diagnostic result correct and clear? (100 points)}
        \begin{itemize}
            \item Reference point: Check whether the following real diagnosis is listed first in the answer:
                \begin{itemize}
                    \item \{real\_diagnosis\}
                \end{itemize}
            \item Scoring:
                \begin{itemize}
                    \item \textbf{100 points}: The first diagnosis given to the patient is the real diagnosis.
                    \item \textbf{50 points}: The real diagnosis is mentioned as one of the possible diagnoses, but not first.
                    \item \textbf{0 points}: The real diagnosis is not mentioned, or the diagnosis is completely incorrect.
                \end{itemize}
        \end{itemize}

    \item \textbf{Does the answer provide the key diagnostic evidence? (100 points)}
        \begin{itemize}
            \item Reference point: Check whether the answer contains the following \textbf{reference diagnostic evidences}:
                \begin{itemize}
                    \item \{diagnosis\_evidences\}
                \end{itemize}
            \item Scoring:
                \begin{itemize}
                    \item \textbf{100 points}: All reference diagnostic evidences are given, with nothing missing.
                    \item \textbf{50 points}: The given evidences contain unnecessary content or omissions.
                    \item \textbf{0 points}: No reference diagnostic evidences are given.
                \end{itemize}
        \end{itemize}

    \item \textbf{Does the answer provide the key treatment suggestions? (100 points)}
        \begin{itemize}
            \item Reference point: Check whether the answer contains the following \textbf{reference treatment suggestions}:
                \begin{itemize}
                    \item \{treatment\_suggestions\}
                \end{itemize}
            \item Scoring:
                \begin{itemize}
                    \item \textbf{100 points}: All reference treatment suggestions are given, with nothing missing.
                    \item \textbf{50 points}: The given suggestions contain unnecessary content or omissions.
                    \item \textbf{0 points}: No reference treatment suggestions are given.
                \end{itemize}
        \end{itemize}

    \item \textbf{Does the answer provide the key lifestyle suggestions? (100 points)}
        \begin{itemize}
            \item Reference point: Check whether the answer contains the following \textbf{reference lifestyle suggestions}:
                \begin{itemize}
                    \item \{lifestyle\_suggestions\}
                \end{itemize}
            \item Scoring:
                \begin{itemize}
                    \item \textbf{100 points}: All reference lifestyle suggestions are given, with nothing missing.
                    \item \textbf{50 points}: The given suggestions contain unnecessary content or omissions.
                    \item \textbf{0 points}: No reference lifestyle suggestions are given.
                \end{itemize}
        \end{itemize}
\end{enumerate}

\textbf{Scoring result format}
\begin{quote}
\{
    ``real\_diagnosis\_score'': \{
        ``diagnosis\_in\_response'': ``Diagnoses in the response, listed in the order they appear'',
        ``reason'': ``Basis for scoring, explaining the reason and the score for this item'',
        ``score'': [100, 50, 0]
    \},
    ``diagnosis\_evidences\_score'': \{
        ``reason'': ``Basis for scoring, explaining the reason and the score for this item'',
        ``score'': [100, 50, 0]
    \},
    ``treatment\_suggestions\_score'': \{
        ``reason'': ``Basis for scoring, explaining the reason and the score for this item'',
        ``score'': [100, 50, 0]
    \},
    ``lifestyle\_suggestions\_score'': \{
        ``reason'': ``Basis for scoring, explaining the reason and the score for this item'',
        ``score'': [100, 50, 0]
    \},
\}
\end{quote}
\end{tcolorbox}
    \caption{Prompt for helpfulness. \Eval~scores the accuracy along with helpfulness.  {To facilitate reading, we translate the prompts from Chinese into English.}}
    \label{fig:prompt-help}
\end{figure*}

\begin{figure*}[htbp]
    \centering
    \small
\begin{tcolorbox}[
% verbatim, % <--- Remove this line
center,
title=Prompt for consistency assessment.,
text width=\linewidth,
boxrule = 1.5pt,
boxsep=2mm,
% fontbody=\ttfamily % Optional: to make the text look like verbatim
]
Input a list of disease names and standardize each disease name in the list. Please follow these requirements during the standardization process:

\begin{enumerate}
    \item \textbf{Merge synonyms or near-synonyms}: Merge names that refer to the same disease or concept into a single unified standard name. For example, GERD and gastroesophageal reflux disease should be considered synonyms and standardized as ``gastroesophageal reflux disease''.
    \item \textbf{Maintain the independence of specific lesions}: If certain names are related but refer to specific pathological features or complications, please keep them as independent entities. For example, gastroesophageal reflux disease and reflux esophagitis are related, but the latter is a complication of the former and should remain independent.
    \item \textbf{Ensure simplicity and accuracy}: The standardized names should be as concise and accurate as possible, avoiding ambiguous or overly lengthy expressions.
\end{enumerate}

\textbf{Example input:}
\begin{quote}
raw\_diagnosis\_1 = ``gastroesophageal reflux disease, chronic gastritis'' \\
raw\_diagnosis\_2 = ``GERD'' \\
raw\_diagnosis\_3 = ``esophagitis or chronic gastritis'' \\
raw\_diagnosis\_4 = ``reflux esophagitis'' \\
\end{quote}

\textbf{Example standardized output (in JSON):}
\begin{quote}
\{
  ``diagnosis\_1'': ``gastroesophageal reflux disease'',
  ``diagnosis\_2'': ``gastroesophageal reflux disease'',
  ``diagnosis\_3'': ``esophagitis'',
  ``diagnosis\_4'': ``reflux esophagitis'',
\}
\end{quote}

\textbf{Input disease name list:}
\begin{quote}
\{diagnoses\}
\end{quote}

Please standardize the input disease name list according to the above requirements.

\end{tcolorbox}
    \caption{Prompt for consistency assessment.  {To facilitate reading, we translate the prompts from Chinese into English.}}
    \label{fig:prompt-cons}
\end{figure*}

\end{document}